\documentclass[10pt,journal,compsoc]{IEEEtran}
\usepackage{graphicx}
\ifCLASSOPTIONcompsoc
  \usepackage[nocompress]{cite}
\else
  \usepackage{cite}
\fi

\usepackage[utf8]{inputenc}
\usepackage{amssymb}
\usepackage{amsmath}
\usepackage{enumerate}
\usepackage{algorithm}
\usepackage[noend]{algpseudocode}
\usepackage{tabularx}
\usepackage{mathtools}
\usepackage{graphicx}
\usepackage{graphicx}
\usepackage[caption=false]{subfig}
\usepackage{stackengine}[2013-10-15]
\usepackage{scalerel}

    \newtheorem{theorem}{Theorem}[section]
    
    \newtheorem{definition}[theorem]{Definition}

    \newcommand{\qed}{\nobreak \ifvmode \relax \else
          \ifdim\lastskip<1.5em \hskip-\lastskip
          \hskip1.5em plus0em minus0.5em \fi \nobreak
          \vrule height0.75em width0.5em depth0.25em\fi}

\graphicspath{{img/}}

\newcommand\DTW{\text{DTW}}

\newcommand{\argmin}{\arg\!\min}

\begin{document}

\title{Times series averaging from a probabilistic interpretation of time-elastic kernel 
}


\author{Pierre-Francois Marteau,~\IEEEmembership{Member,~IEEE,}
\IEEEcompsocitemizethanks{\IEEEcompsocthanksitem P.-F. Marteau is with UMR CNRS IRISA, Université de Bretagne Sud, F-56000 Vannes, France.}\protect\\
E-mail: see http://people.irisa.fr/Pierre-Francois.Marteau/
\thanks{}}

\begin{abstract}
In the light of regularized dynamic time warping kernels, this paper re-considers the concept of time elastic centroid (TEC) for a set of time series. From this perspective, we show that TEC can be readily addressed as a preimage problem. However, this non-convex problem is ill-posed,  and obtaining a sub-optimal solution may involve heavy computational costs, especially for long time series. We then derive two new algorithms based on a probabilistic interpretation of kernel alignment matrices that expresses the result in terms of probabilistic distributions over sets of alignment paths. The first algorithm is an agglomerative iterative heuristic procedure inspired from a state-of-the-art DTW barycentre averaging algorithm. The second proposed algorithm uses a  progressive agglomerative heuristic method to perform classical averaging of the aligned samples but also averages the times of occurrence of the aligned samples. By comparing classification accuracies for 45 time series datasets obtained by first nearest centroid/medoid classifiers we show that: i) centroid-based approaches significantly outperform medoid-based approaches, ii) for the considered datasets, the second algorithm which combines averaging in the sample space and along the time axes, emerges as the most significantly robust heuristic model for time-elastic averaging with a promising noise reduction capability.

\end{abstract}


\maketitle

\section{Introduction}
\label{intro}
Since Maurice Fr\'{e}chet's pioneering work \cite{frechet1906} in the early 1900s, \textit{time-elastic} matching of time series or symbolic sequences has attracted much attention from the scientific community in numerous fields such as information indexing and retrieval, pattern analysis, extraction and recognition, data mining, etc. This approach has impacted a very wide spectrum of applications relating to a multitude of socio-economic issues such as the environment, industry, health, energy, defense and so on.

Among other time elastic measures, Dynamic Time Warping (DTW) was widely popularized during the 1970s with the advent of speech recognition systems
\cite{VelichkoZagoruyko1970}, \cite{SakoeChiba1971} and numerous variants that have since been proposed to match time series with a certain degree of time distortion tolerance.

The main issue addressed here is time series or shape averaging in the context of a time elastic distance. This is a long-standing issue that is currently becoming increasingly prevalent; it is relevant for summarizing subsets of time series, defining significant prototypes, identifying outliers, performing data mining tasks (mainly exploratory data analysis such as clustering) and speeding up classification, as well as regression or data analysis processes in a big data context.

In this paper, we specifically tackle the question of averaging subsets of time series, not from considering the DTW measure itself as has been already largely explored, but from the perspective of the so-called regularized DTW kernel (KDTW) that ensures positive definiteness. From this new perspective, the estimation of a time series average or centroid can be readily addressed as a preimage (inverse) problem.  However, this approach has some theoretical and practical limitation that are discussed in the following sections. A more promising direct approach is developed here, which is based on a probabilistic interpretation of kernel alignment matrices, allowing a precise definition of the average of a pair of time series from the expected value of local alignments of samples. The tests carried out so far demonstrate the robustness and the efficiency of this approach comparison to the state-of-the art approach.

The structure of this paper is as follows: after an introduction, the second section summarizes the most relevant related studies on time series averaging as well as DTW kernelization. In the third section, we show how the estimation of a time-elastic centroid can be addressed as a preimage problem in the context of the DTW regularized kernel (KDTW). In the fourth section, we derive a probabilistic interpretation from the kernel alignment matrices evaluated on a pair of time series. In the fifth section, we define the average of a pair of time series, and based on this pairwise averaging procedure, we propose two sub-optimal algorithms designed for the averaging of any subset of time series. 

\section{Related works}
\label{sec:RelatedWorks}
Time series averaging in the context of (multiple) time elastic distance alignments has been mainly addressed in the scope of the Dynamic Time Warping (DTW) measure \cite{VelichkoZagoruyko1970}, \cite{SakoeChiba1971}. Although other time elastic distance measures such as the Edit Distance With Real Penalty (ERP) \cite{Chen04ERP} or the Time Warp Edit Distance (TWED) \cite{Marteau09TWED} could be considered instead, without loss of generality, we remain focused throughout this paper on DTW and its kernelization. 

\subsection{DTW and time elastic centroid of a pair of time series}
A classical formulation of DTW can be given as follows. If $d$ is a fixed positive integer, we define a time series of length $T$ as a multidimensional sequence $v=v(i)$, such that, $\forall i \in \{1, ..,T\}$,  $v(i) \in \mathbb{R}^d$.\\

\begin{definition}
\label{def:alignmentPath}
    If $u$ and $v$ are two time series with respective lengths $T_1$ and $T_2$, an {\it alignment path} $\pi = (\pi_k)$ of length $p=|\pi|$ between $u$ and $u$ is represented by a sequence
    \[
        \pi : \{1, \ldots, p\} \rightarrow \{1, \ldots, T_1\} \times \{1, \ldots, T_2\}
    \]
    such that $\pi_1 = (1, 1)$, $\pi_p = (T_1, T_2)$, and (using the
    notation $\pi_k = (i_k, j_k)$, for all $k \in \{1, \ldots, p-1\}$,
    $\pi_{k+1} = (i_{k+1}, j_{k+1}) \in \{(i_k + 1, j_k),\linebreak[1](i_k, j_k + 1),\linebreak[1](i_k + 1, j_k + 1) \}$.\\

We define $\forall k$ $\pi_{k}(1)=i_k$ and $\pi_{k}(2)=j_k$, as the index access functions at step $k$ of the mapped elements in the pair of aligned time series.\\
\end{definition}

    In other words, a warping path defines a way to travel along both time series simultaneously from beginning to end; it cannot skip a point, but it can advance one time step along one series without advancing along the other, thereby justifying the term \textit{time-warping}.

    If $\delta$ is a distance on $\mathbb{R}^d$, the global {\it cost}
    of a warping path $\pi$ is the sum of distances (or squared distances or local costs) between pairwise elements of the two time series along $\pi$, i.e.:
    \[
        \text{cost}(\pi) = \sum_{(i_k,j_k) \in \pi} \delta(v_{i_k}, w_{j_k})
    \]
    A common choice of distance on $\mathbb{R}^d$ is the one generated by the
    $L^2$ norm:
    \[
      \delta(x, y) = \|x - y\|_2^2 =\sum_{l=1}^d (x_l - y_l)^2.
    \]

\begin{definition}
\label{dtw}
    For a finite time series, any warping path has a finite length, and thus the number of existing warping paths is finite. Hence, there exists at least one path $\pi^*$ whose cost is minimal, so we can define $\text{DTW}(u, v)$ as the minimal cost taken over all existing warping paths. Hence
\begin{equation}
\label{eq:dtw}
      \text{DTW}(u, v) = \underset{\pi}{\min} \text{ cost}(\pi(u,v))=\text{cost}(\pi^*(u,v)).
\end{equation}

\end{definition}

\begin{definition}
\label{pairTScentroid}
From the DTW measure, it is straightforward to define the time elastic centroid $c(u,v)$ of a pair of time series $u$ and $v$ as the time series $(c_k)$ whose elements are $c_k=\textsl{Centroid}(u(\pi^*_k(1)), v(\pi^*_k(2))$, $\forall k \in {1, \cdots, |\pi^*|}$, where \textsl{Centroid} corresponds to the usual definition in Euclidean space.\\
\end{definition}

  \begin{figure*}[!ht]
    \subfloat[Progressive agglomeration\label{fig:hac-iter-a}]{%
      	\fbox{\includegraphics[scale=.4]{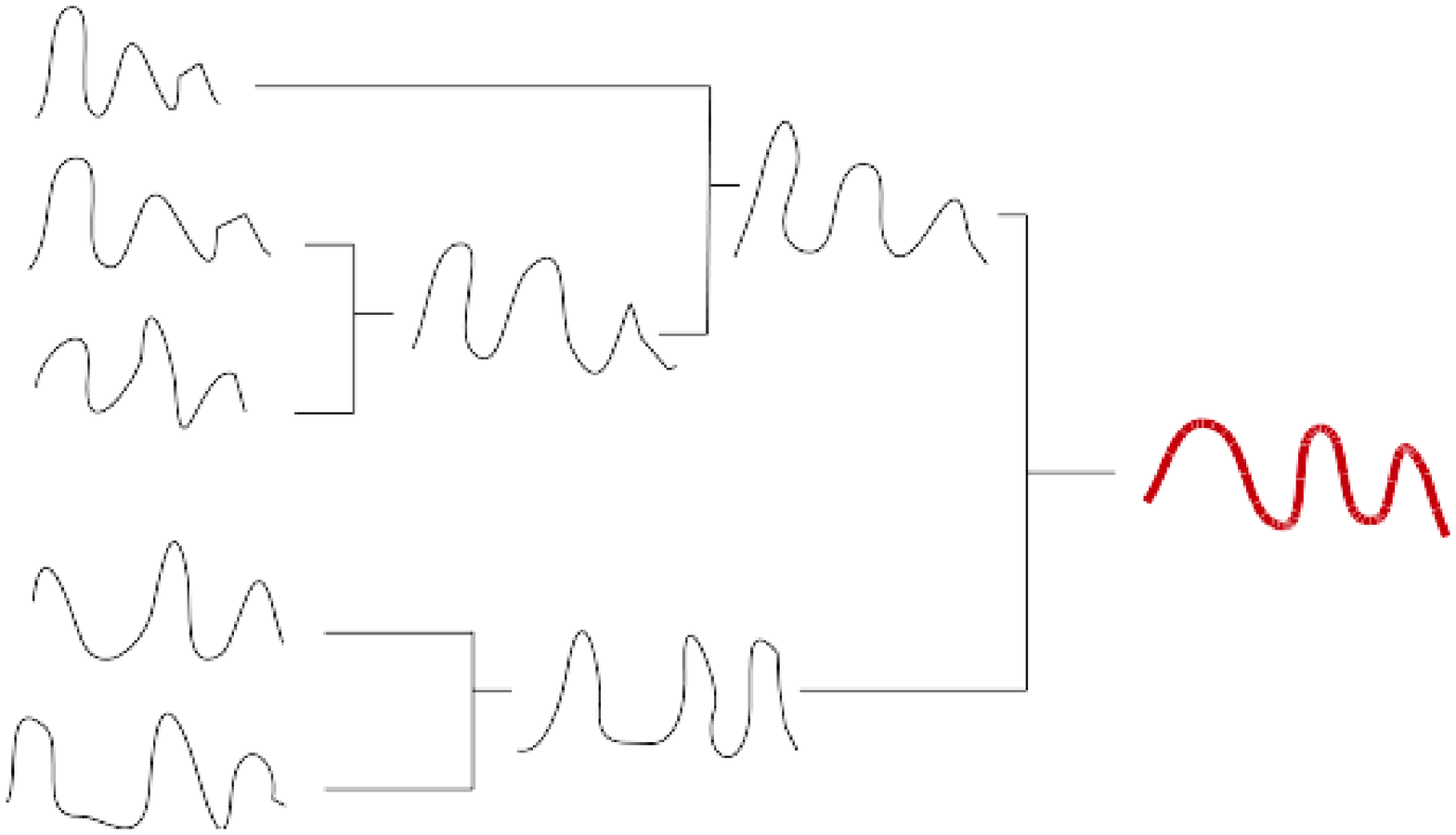}}
    }
    \hfill
    \subfloat[Iterative agglomeraation with refinement\label{fig:hac-iter-b}]{%
      	\fbox{\includegraphics[scale=.4]{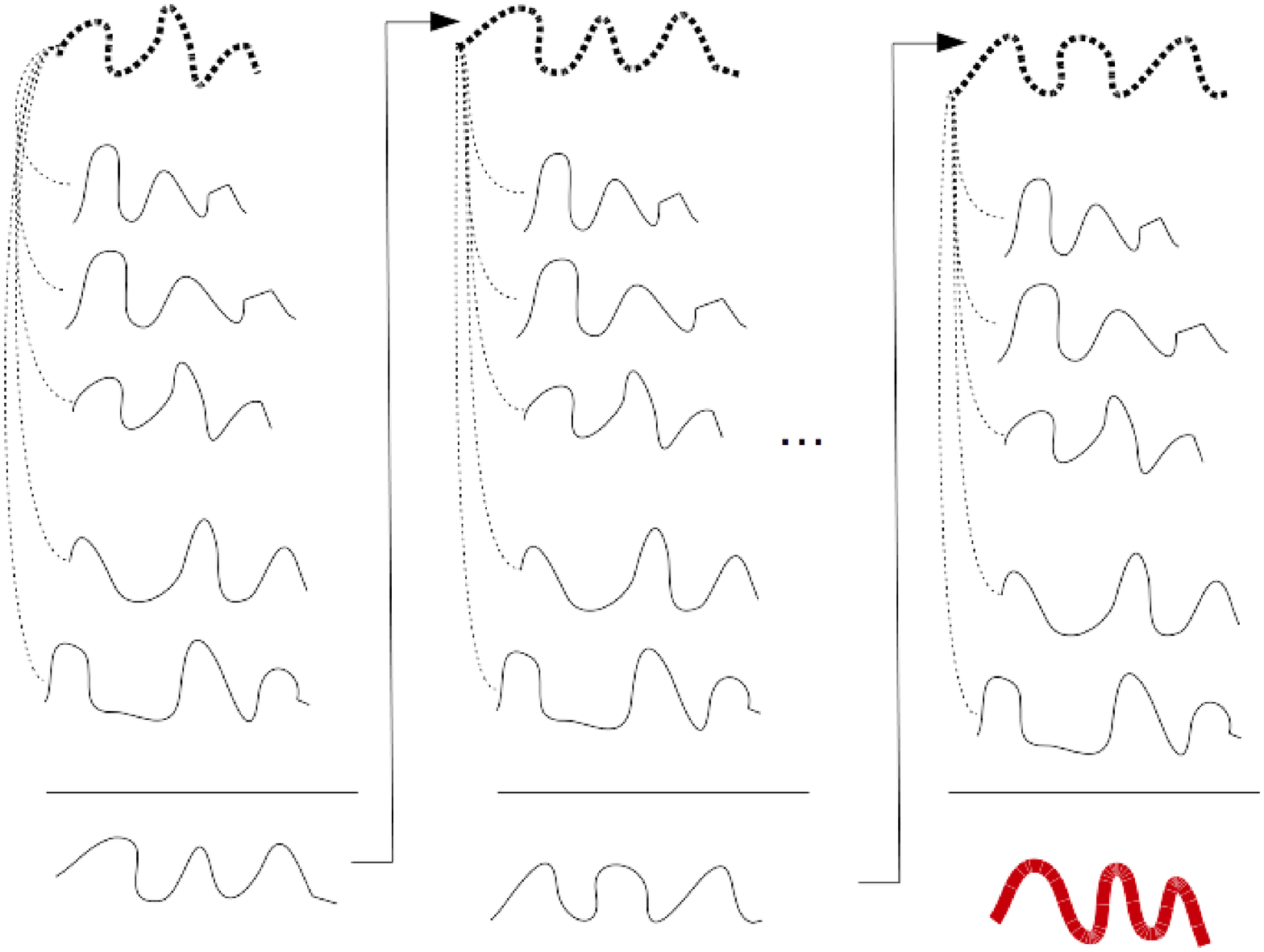}}
    }
    \caption{Progressive hierarchical with similar first agglomeration (left) v.s. iterative agglomeration (right) strategies. Final centroid approximations are presented in red bold color. Temporary estimations are presented using a bold dotted black line}
    \label{fig:hac-iter}
  \end{figure*}

\subsection{Time elastic centroid of a set of time series}
A single alignment path is required to calculate the time elastic centroid of a pair of time series (Def. \ref{pairTScentroid}). However, multiple path alignments need to be considered to evaluate the centroid of a larger set of time series. Multiple alignments have been widely studied in bioinformatics \cite{Fasman1998}, and it has been shown that the computational complexity of determining the optimal alignment of a set of sequences under the sum of all pairs (SP) score scheme is a NP-complete problem \cite{WangJ1994} \cite{Just99}. The time and space complexity of this problem is $O(L^k)$, where $k$ is the number of sequences in the set and $L$ is the length of the sequences when using dynamic programming to search for an optimal solution \cite{Carrillo1988}. This latter result applies to the estimation of the time elastic centroid of a set of $k$ time series with respect to the DTW measure. Since the search for an optimal solution becomes rapidly intractable with increasing $k$, sub-optimal heuristic solutions have been subsequently proposed, most of them falling into one of the following three categories.

\subsubsection{Progressive heuristics}
Progressive heuristic methods estimate the time elastic centroid of a set of $k$ time series by combining pairwise centroids (Def. \ref{pairTScentroid}). This kind of approach constructs a binary tree whose leaves correspond to the time series of the data set, and whose nodes correspond to the calculation of a local pairwise centroid, such that, when the tree is complete, the root is associated with the estimated data set centroid. The proposed strategies differ in the way the tree is constructed. One popular approach consists of providing a random order for the leaves, and then constructing the binary tree up to the root using this ordering \cite{Gupta1996}. Another approach involves constructing a dendrogram (a hierarchical ascendant clustering) from the data set and then using this dendrogram to calculate pairwise centroids starting with the closest pairs of time series and progressively aggregating series that are farther away\cite{Niennattrakul2009} as illustrated on the left of Fig. \ref{fig:hac-iter}. Note that these heuristic methods are entirely based on the calculation of a pairwise centroid, so they do not explicitly require the evaluation of a DTW centroid for more than two time series. Their degree of complexity varies linearly with the number of time series in the data set. 

\subsubsection{Iterative heuristics}
Iterative heuristics are based on an iterated three-step process. For a given temporary centroid candidate, the first step consists of calculating the inertia, i.e. the sum of the DTW distances between the temporary centroid and each time series in the data set. The second step evaluates the best pairwise alignment with the temporary centroid for each time series $u_j(i)$ in the data set ($j \in \{1 \cdots n\}$). A new time series  $\tilde{u}_j(i)$ is thus constructed that contains all the samples of time series $u_j(i)$,  but with time being stretched or compressed according to the best alignment path. The third step consists of producing a new temporary centroid candidate $c(i)$ from the set $\{\tilde{u}_j(i)\}$ by successively averaging (in the sense of the Euclidean centroid), the samples at every timestamp $i$ of the $\tilde{u}_j(i)$ time series. Basically, $c(i) = \sum_{j=1..n_i} \tilde{u}_j(i).1\!\!1(i,j)/\sum_{j=1..n_i} 1\!\!1(i,j)$), where $1\!\!1(i,j)$ is an indicator function equal to $1$ if time series $\tilde{u}_j$ is defined for timestamp $i$, but which is otherwise $0$. 

Thus, the new centroid candidate replaces the previous one and the process is iterated until the inertia is no longer reduced or the maximum number of iterations is reached. Generally, the first temporary centroid candidate is taken as the DTW medoid of the considered data set. This process is illustrated on the right of Fig. \ref{fig:hac-iter}. The three steps of this heuristic method were first proposed in \cite{Abdulla2003}. The iterative aspect of this heuristic approach was initially introduced by \cite{Hautamaki2008} and refined by \cite{Petitjean2011}. Note that, in contrast to the progressive method, this kind of approach needs to evaluate, at each iteration, all the alignments with the current centroid candidate. The complexity of the iterative approach is higher than the progressive approach, the extra computational cost being linear with the number of iterations. More sophisticated approaches have been proposed to escape some local minima. \cite{Petitjean2012} have evaluated a genetic algorithm for managing a population of centroid candidates,  thus improving with some success the straightforward iterative heuristic methods.    

\subsubsection{Optimization approaches} 
Given the entire set of time series $\mathbb{S}$ and a subset of $n$ time series $S=\{u_j\}_{j=1 \cdots n} \subseteq \mathbb{S}$, optimization approaches attempt to estimate the centroid of $S$ from the definition of an optimization problem, which is generally expressed by Eq. \ref{eq.opt} given below: 
\begin{equation}
\label{eq.opt}
c = \argmin_{s \in \mathcal{S}} \sum_{j=1}^n \DTW(s,u_j) 
\end{equation}
To our knowledge, the first attempt to use this kind of direct approach for the estimation of time elastic centroid estimation was recently described in \cite{Soheily-Khal:2015}.

These authors (op.cit.) derived a solution of their original non-convex constrained optimization problem, by integrating a temporal weighting of local sample alignments to highlight the temporal region of interest in a time series data set, thus penalizing the other temporal regions. Two time elastic measures were specifically addressed: i) a dynamic time warping measure between a time series and a weighted time series (representing the centroid estimate) and ii) an (indefinite) kernel DTW called DTAK \cite{Shimodaira2002}. Their results are very promising: although the number of parameters to optimize is linear with the size and the dimensionality of the time series, the two steps gradient-based optimization process they derived is very computationally efficient and shown to outperform the state of the art approaches on some challenging scalar and multivariate data sets. However, as numerous local \textit{optima} exist in practice, the method is not guaranteed to converge toward the best possible centroid, which is anyway the case in all other approaches.

\subsection{Discussion and motivation}
According to the state of the art in time elastic centroid estimation, an exact centroid, if it exists, can be calculated by solving a NP-complete problem whose complexity is exponential with the number of time series to be averaged. Heuristic methods with increasing time complexity have been proposed since the early 2000s. Simple pairwise progressive aggregation is a less complex approach, but which suffers from its dependence on initial conditions. Iterative aggregation is reputed to be more efficient, but entails a higher computational cost. It could be combined with ensemble methods or soft optimization such as genetic algorithms. The non-convex optimization approach has the merit of directly addressing the mathematical formulation of the centroid problem in a time elastic distance context. This approach nevertheless involves a higher complexity and must deal with a relatively large set of parameters to be optimized (the weights and the sample of the centroid). Its scalability could be questioned, specifically for high dimensional multivariate time series.   

It should also be mentioned that some criticisms of these heuristic methods have been made in \cite{Niennattrakul2007}. Among other drawbacks, the fact that DTW is not a metric (the triangle inequality is not satisfied) could explain the occurrence of unwanted behaviour such as centroid drift outside the time series cluster to be averaged. We should also be borne in mind that keeping a single best alignment (even though several may exist, without mentioning the \textit{good} ones) can increase the dependence of the solution on the initial conditions. It may also increase the aggregating order of the time series proposed by the chosen method, or potentially enhance the convergence rate.    

In this study, we do not directly address the issue of time elastic centroid estimation from the DTW perspective, but rather from the point of view of the regularized dynamic time warping kernel (KDTW) \cite{MarteauGibet2014}. This perspective allows us to consider centroid estimation as a preimage problem, which is in itself another optimization perspective. More importantly, the KDTW alignment matrices can be used to derive a probabilistic interpretation of the pairwise alignment of time series. This leads us to propose a robust interpolation scheme jointly along the time axis and in the sample space. We do not claim that using KDTW and its probabilistic interpretation can solve all or even any of the fundamental questions raised earlier: since the problem tackled here is NP-complete, an exact solution requires exponentially complex computations and any heuristic method must handle numerous local minima. Our aim is to throw some new light on the problem as well as obtain new quantitative results showing, in this difficult context, that the proposed alternative approach is worth considering. 

\subsection{Time elastic kernels and their regularization}    
\label{sec:DTW}
\textbf{Dynamic Time Warping} (DTW), \cite{VelichkoZagoruyko1970}, \cite{SakoeChiba1971} as defined in Eq.\ref{eq:dtw} can be recursively evaluated as 

\begin{eqnarray}
\label{Eq.dtw2}
 d_{dtw}(X_p,Y_q)&= &d_{E}^{2}(x(p),y(q))  \\
  &+&\text{Min} 
   \left\{
   \begin{array}{ll}
     d_{dtw}(X_{p-1},Y_{q}) & sup \nonumber\\ 
     d_{dtw}(X_{p-1},Y_{q-1}) & sub	\nonumber	 \\
     d_{dtw}(X_{p},Y_{q-1}) & ins \nonumber \\
   \end{array}
   \right.
\end{eqnarray}
where $d_{E}(x(p),y(q)$ is the Euclidean distance (eventually, the square of the Euclidean distance) defined on $\mathbb{R}^k$ between the two positions/?points in sequences $X$ and $Y$ taken at times $p$ and $q$, respectively. 
 
Apart from the fact that the triangular inequality does not hold for the DTW distance measure, it is furthermore not possible to define a positive definite kernel directly from this distance. Hence, the optimization problem, which is inherent to the learning of a kernel machine, is no longer quadratic and, at least for some tasks, could be a source of limitation.\\

\textbf{Regularized DTW}: recent studies  \cite{CuturiVert2007},  \cite{MarteauGibet2014} lead us to propose new guidelines to ensure that kernels constructed from elastic measures such as DTW are positive definite. A simple instance of such a regularized kernel, derived from \cite{MarteauGibet2014}, can be expressed in the following form, which makes use of two recursive terms:

\begin{align}
\label{Eq.KDTW}
\begin{array}{ll}
\textsc{KDTW} (X_{p},Y_{q})=K^{xy}_{dtw}(X_{p}, Y_{q})+K^{xx}_{dtw}(X_{p},Y_{q}) \\
\\
K^{xy}_{dtw}(X_{p},Y_{q}) = \frac{1}{3}e^{-\nu d_{E}^{2}(x(p),y(q))}  \\
   \sum \left\{
   \begin{array}{ll}
    h(p-1,q)K^{xy}_{dtw}(X_{p-1},Y_{q}) \\
   h(p-1,q-1) K^{xy}_{dtw}(X_{p-1},Y_{q-1})  \\
    h(p,q-1)K^{xy}_{dtw}(X_{p},Y_{q-1}) \\
   \end{array}
   \right.\\
\\
   K^{xx}_{dtw}(X_{p},Y_{q}) = \frac{1}{3} \\
   \sum \left\{
   \begin{array}{ll}
    h(p-1,q) K^{xx}_{dtw}(X_{p-1},Y_{q})e^{-\nu d_{E}^{2}(x(p),y(p))}  \\
    \Delta_{p,q} h(p,q)K^{xx}_{dtw}(X_{p-1},Y_{q-1})e^{-\nu d_{E}^{2}(x(p),y(q))}   \\
    h(p,q-1)K^{xx}_{dtw}(X_{p},Y_{q-1})e^{-\nu d_{E}^{2}(x(q),y(q))} \\
   \end{array}
   \right.\\
  \end{array}
\end{align}
where $\Delta_{p,q}$ is the Kronecker symbol, $\nu \in \mathbb{R}^{+}$ is a \textit{stiffness} parameter which weights the local contributions, i.e. the distances between locally aligned positions, and $d_E(.,.)$ is a distance defined on $\mathbb{R}^{k}$. 

The initialization is simply $K^{xy}_{dtw}(X_{0},Y_{0}) = K^{xx}_{dtw} (X_{0},Y_{0}) = 1$.\\

The main idea behind this regularization is to replace the operators $\min$ and $\max$ (which prevent symmetrization of the kernel) by a summation operator ($\sum$). This allows us to consider the best possible alignment, as well as all the best (or nearly the best) paths by summing their overall cost. The parameter $\nu$ is used to check what is termed as nearly-the-best alignment, thus penalizing alignments that are too far away from the optimal ones. This parameter can be easily optimized through a cross-validation. \\

\section{KDTW centroid as a preimage problem}

In this section, we tackle the centroid estimation question from a \textit{kernelized centroid} point of view, the kernel of interest being KDTW.  \\

The Moore-Aronszajn theorem \cite{aronszajn1950} establishes that a reproducing kernel Hilbert space (RKHS) exists uniquely for every positive definite kernel and \textit{vice-versa}. Let $\mathcal{H}$ be the RKHS associated to kernel $\kappa$ defined on a set
 $\mathcal{X}$, and let $\langle.,.\rangle_{\mathcal{H}}$ be the inner product defined on
  $\mathcal{H}$. In addition, the representer property of the evaluation functional in $\mathcal{H}$ is expressed as: for any $\psi \in \mathcal{H}$ and any $x_j \in \mathcal{X}$, $\psi(x_j) = \langle \psi(.), \kappa(.,x_j)\rangle_{\mathcal{H}}$. 
  
  Denoting  $\phi(.)$ as the map that assigns the kernel function $\kappa(.,x)$ to each input $x \in \mathcal{X}$, the reproducing property of the kernel implies that for any 
$(x_i, x_j) \in \mathcal{X}^2$, $\kappa(x_i, x_j)=\langle\phi(x_i),\phi(x_j)\rangle_{\mathcal{H}}$.

Furthermore, $D_\mathcal{H}(x_i,x_j)^2=||\phi(x_i) - \phi(x_j)||_\mathcal{H}^2=\langle\phi(x_i),\phi(x_i)\rangle_{\mathcal{H}} + \langle\phi(x_j),\phi(x_j)\rangle_{\mathcal{H}} -2.\langle\phi(x_i),\phi(x_)\rangle_{\mathcal{H}}$ is the generalization of the squared Euclidean distance defined in the feature space $\mathcal{H}$, which can be expressed in kernel terms as: $D_\mathcal{H}(x_i,x_j)^2=\kappa(x_i,x_i)^2+\kappa(x_j,x_j)^2 -2.\kappa(x_i,x_j)$ (the so-called kernel trick). 

Finally, the representer theorem \cite{Scholkopf2001} states that any function $\varphi(.)^*$ of a RKHS   $\mathcal{H}$ minimizing a regularized cost functional of the form: 
\[
\sum_{i=1}^n \mathbf{J}(\varphi(x_i), y_i) + g(||\varphi||_\mathcal{H}^2)
\]
with predicted output $\varphi(x_i)$ for input $x_i$ and desired output $y_j$, where $g(.)$ is a strictly monotonically increasing function on $\mathbb{R}^+$-, is equivalent to a kernel expansion expressed in terms of available data ($\{(x_i,y_i)\}$) 

\begin{equation}
\label{eq:span}
\varphi^*(.)=\sum_{i=1}^n \gamma_i \kappa(x_i,.) \text{, where } \forall i, \gamma_i \in \mathbb{R}.
\end{equation}

Hence, a direct definition of the  kernelized centroid of the set $\{x_i, i=1..n\}$  expressed in the RKHS $\mathcal{H}$ feature space associated with kernel $\kappa$ can be written as:

\begin{align}
\label{eq:cntrImage}
\varphi^*(.) &= \argmin_{\varphi(.) \in \mathcal{H}} \sum_{i=1}^n ||\varphi(.) - \kappa(.,x_i)||_{\mathcal{H}}^2  \\
   &= \argmin_{\varphi(.) \in \mathcal{H}} n\cdot||\varphi(.)||_{\mathcal{H}}^2 -2\cdot\sum_{j=1}^n \langle\varphi(.),\kappa(., x_j)\rangle_{\mathcal{H}} \nonumber
\end{align}

The representer theorem applies and thus $\varphi^*(.)$ takes the form given in Eq. \ref{eq:span}, which allows us to rewrite Eq. \ref{eq:cntrImage} as follows:

\begin{align}
\label{eq:cntrImage2}
\varphi^*(.) = &\argmin_{\{\lambda_i \}_{i=1 \cdots n}} \sum_{i=1}^n \sum_{j=1}^n \gamma_i \gamma_j \kappa(x_i, x_j) \nonumber \\
   & -2\cdot \sum_{i=1}^n \sum_{j=1}^n \gamma_j \kappa(x_i, x_j)
\end{align}

\begin{figure}[t!]
\centering
\begin{tabular}{ccc}
    	\fbox{\includegraphics[scale=.33]{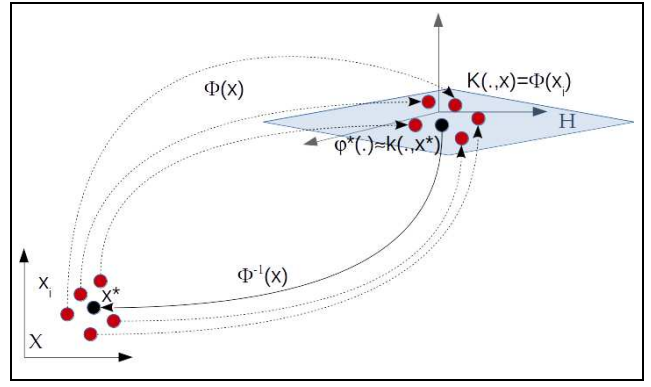}}
\end{tabular}
\caption{Centroid estimation viewed as a preimage problem.}
\label{fig:preimage}
\end{figure}

Unfortunately, if the kernelized centroid is related to a well-defined quadratic optimization problem in the RKHS space $\mathcal{H}$ (Eq. \ref{eq:cntrImage2}), it is an ill-posed problem in set $\mathcal{X}$. This is known as the preimage problem, since the pre-image of $\phi(.)^*$ might not exist. Instead, we are seeking the best approximation, namely $x^* \in \mathcal{X}$ whose map $\phi(x^*)=\kappa(.,x^*)$ is as close as possible to $\varphi(.)^*$, as illustrated in Fig.\ref{fig:preimage}. 

Hence, if we remove the term that does depend upon $x$, the optimization problem becomes:

\begin{align}
\label{eq:cntrImage3}
x^* & = \argmin_{x \in \mathcal{X}} n\cdot||\kappa(.,x)||_{\mathcal{H}}^2 -2\cdot\sum_{j=1}^n \langle\kappa(.,x),\kappa(., x_j)\rangle_{\mathcal{H}} \nonumber\\
    &= \argmin_{x \in \mathcal{X}} n\cdot \kappa(x,x) - 2\cdot \sum_{j=1}^n  \kappa(x,x_j)
\end{align}

\begin{figure}[!]
\centering
\begin{tabular}{cc}
    {\includegraphics[scale=.25]{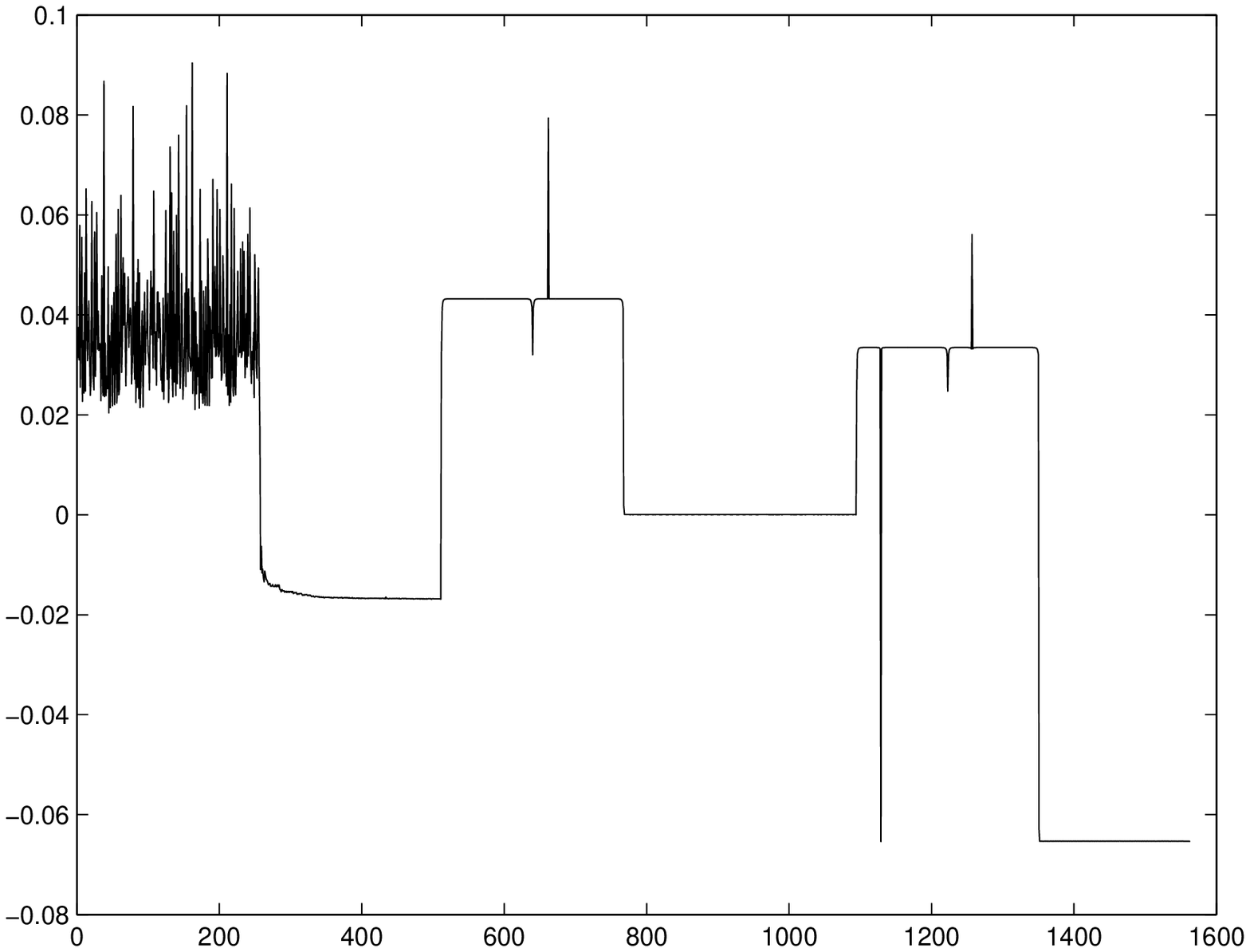}}
    	{\includegraphics[scale=.25]{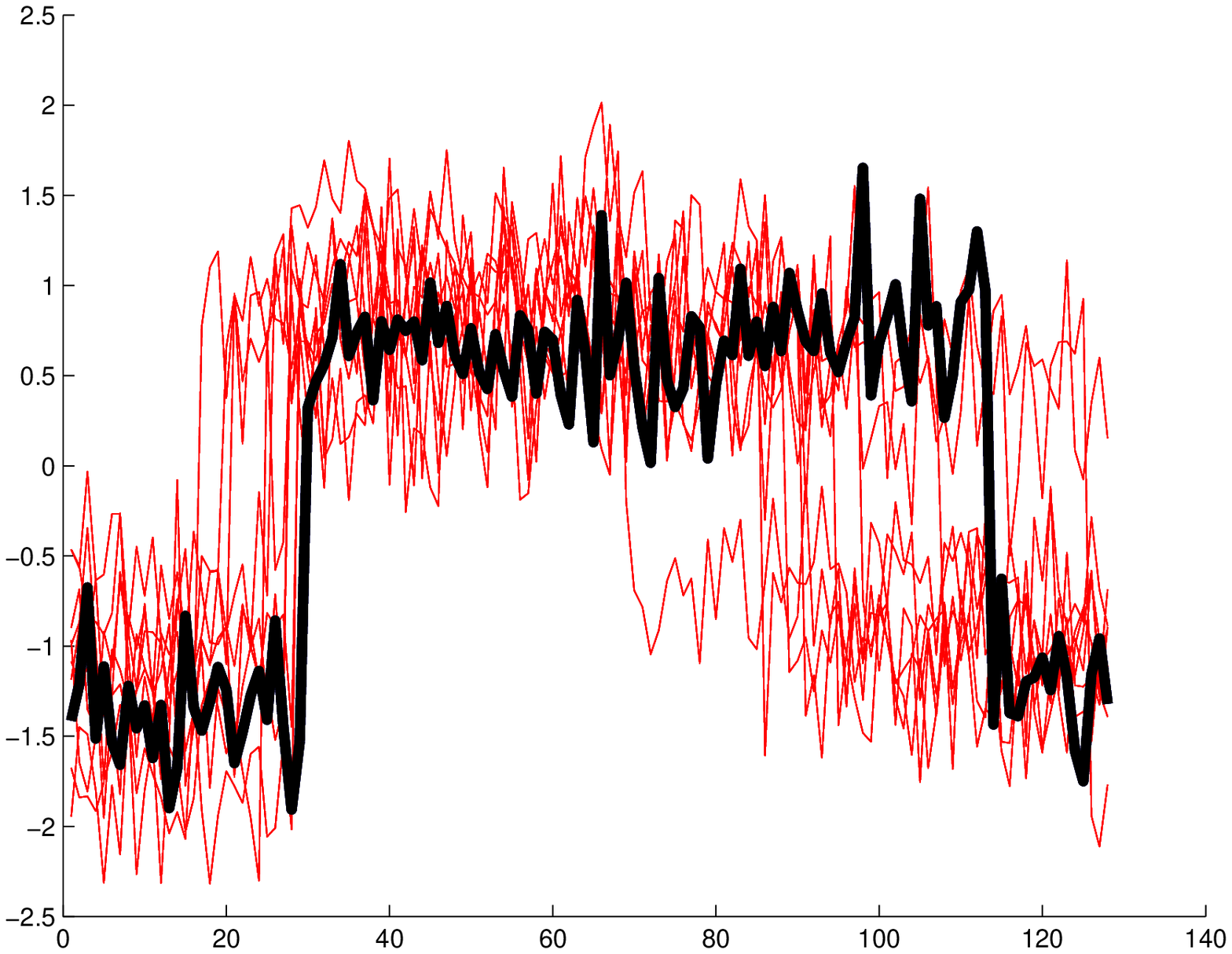}}\\
    {\includegraphics[scale=.25]{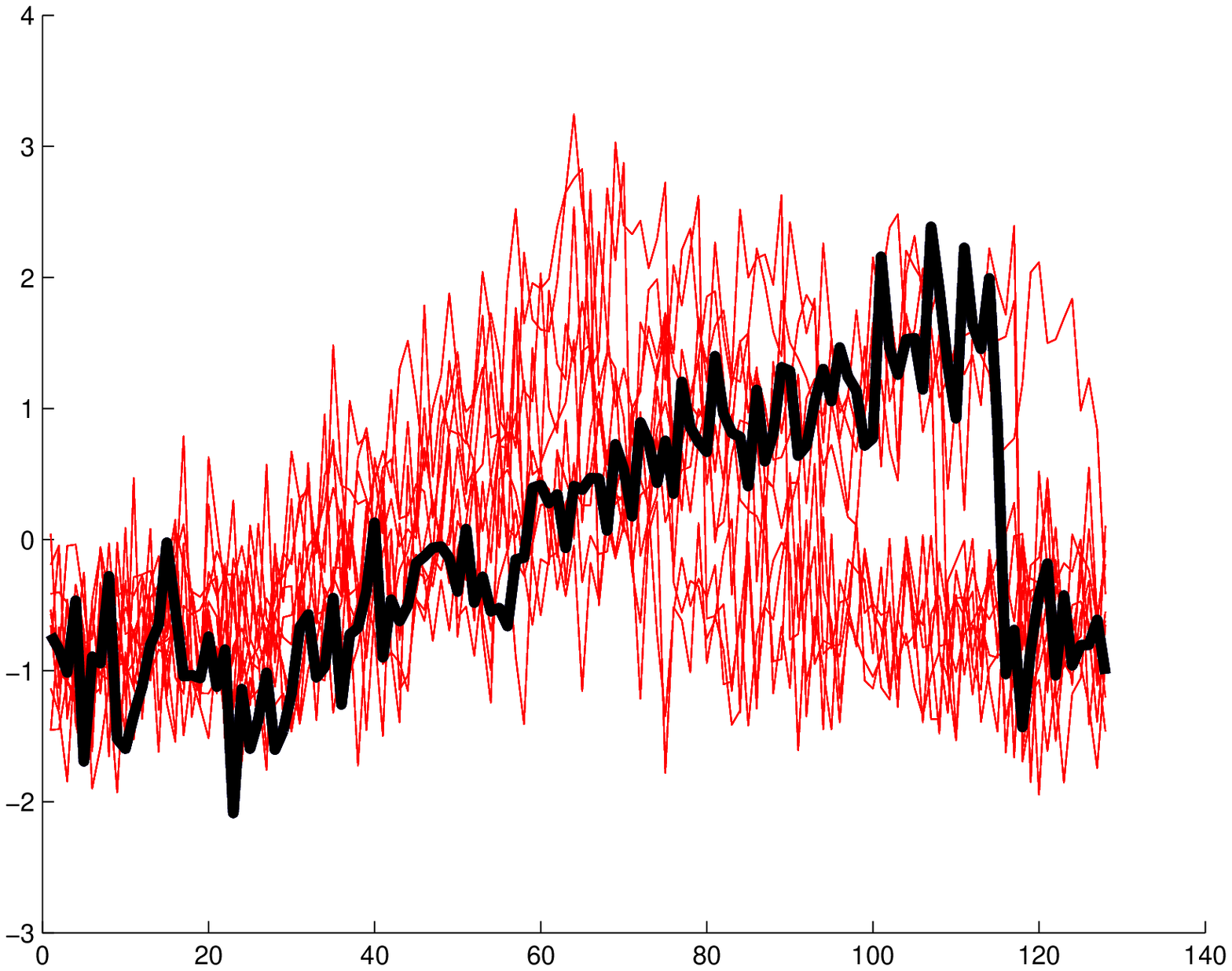}}
    	{\includegraphics[scale=.25]{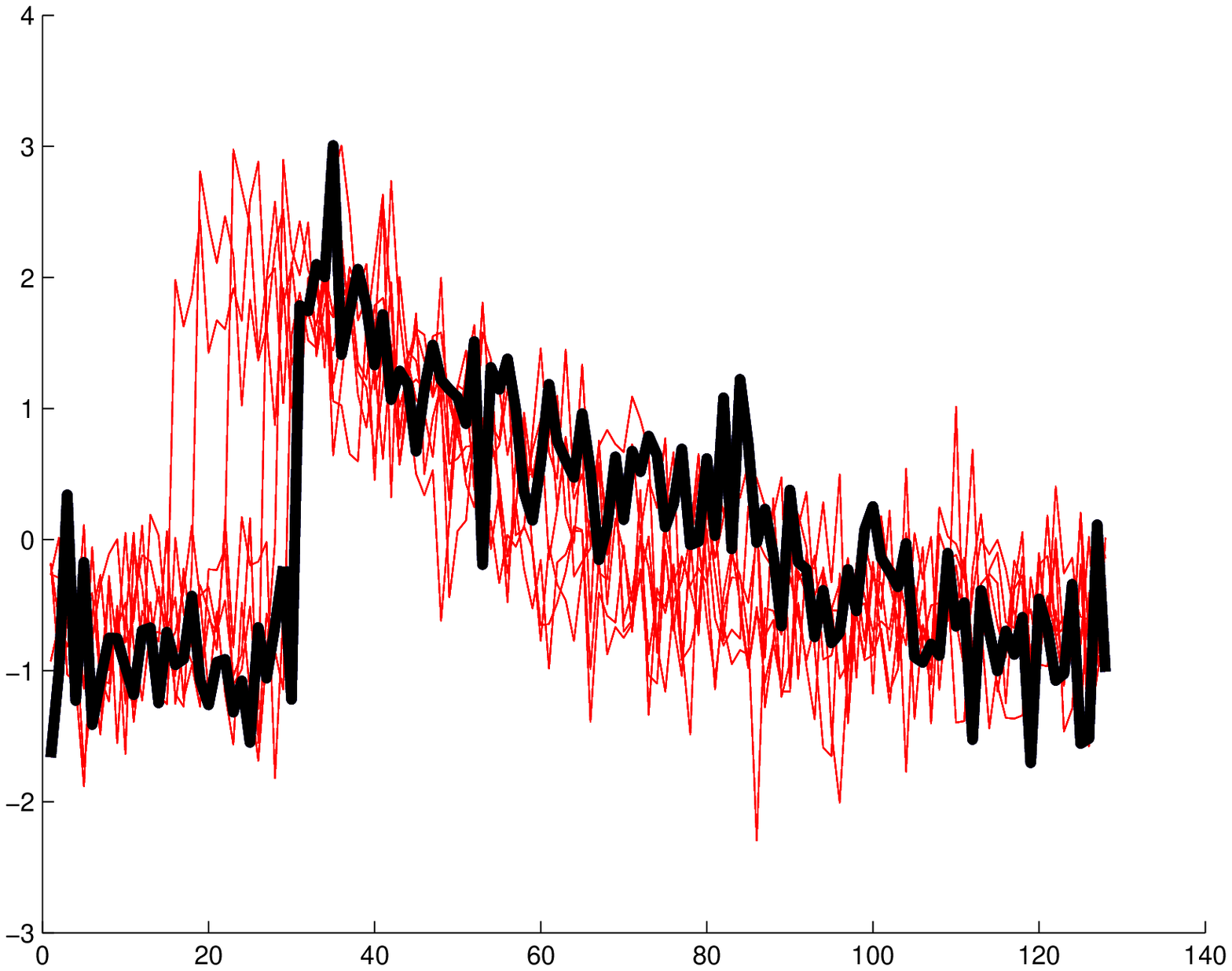}}
\end{tabular}
\caption{Centroid estimation for the three categories contained in the CBF dataset as a solution of the preimage problem. In bold, the centroid time series; in light red, the time series of the averaged dataset. At top left of figure, the value of the minimized functional is plotted on a log-scale versus the iteration index.}
\label{fig:preimageCBF}
\end{figure}

\begin{figure}[!]
\centering
\begin{tabular}{cc}
    	{\includegraphics[scale=.25]{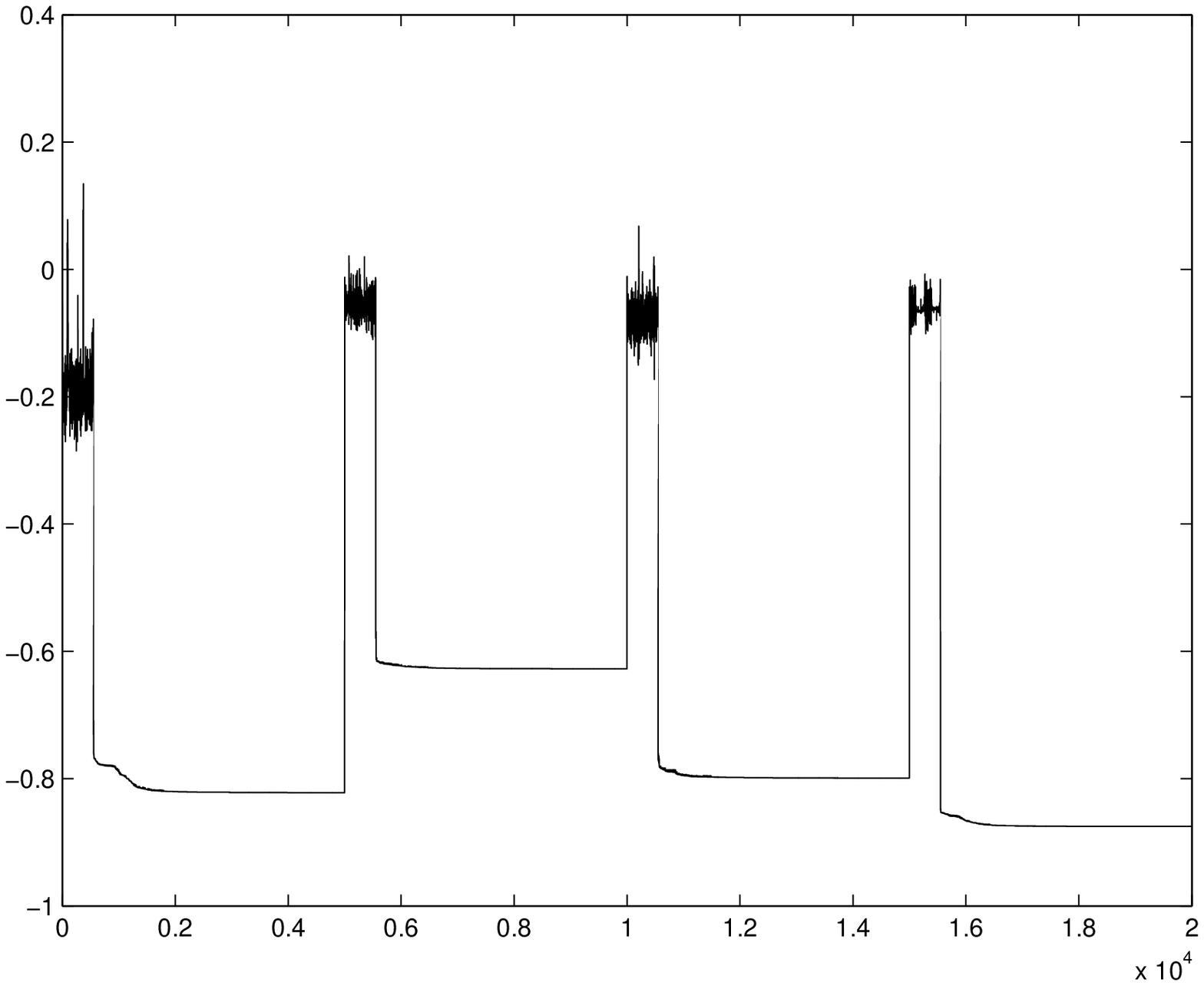}}
    	{\includegraphics[scale=.25]{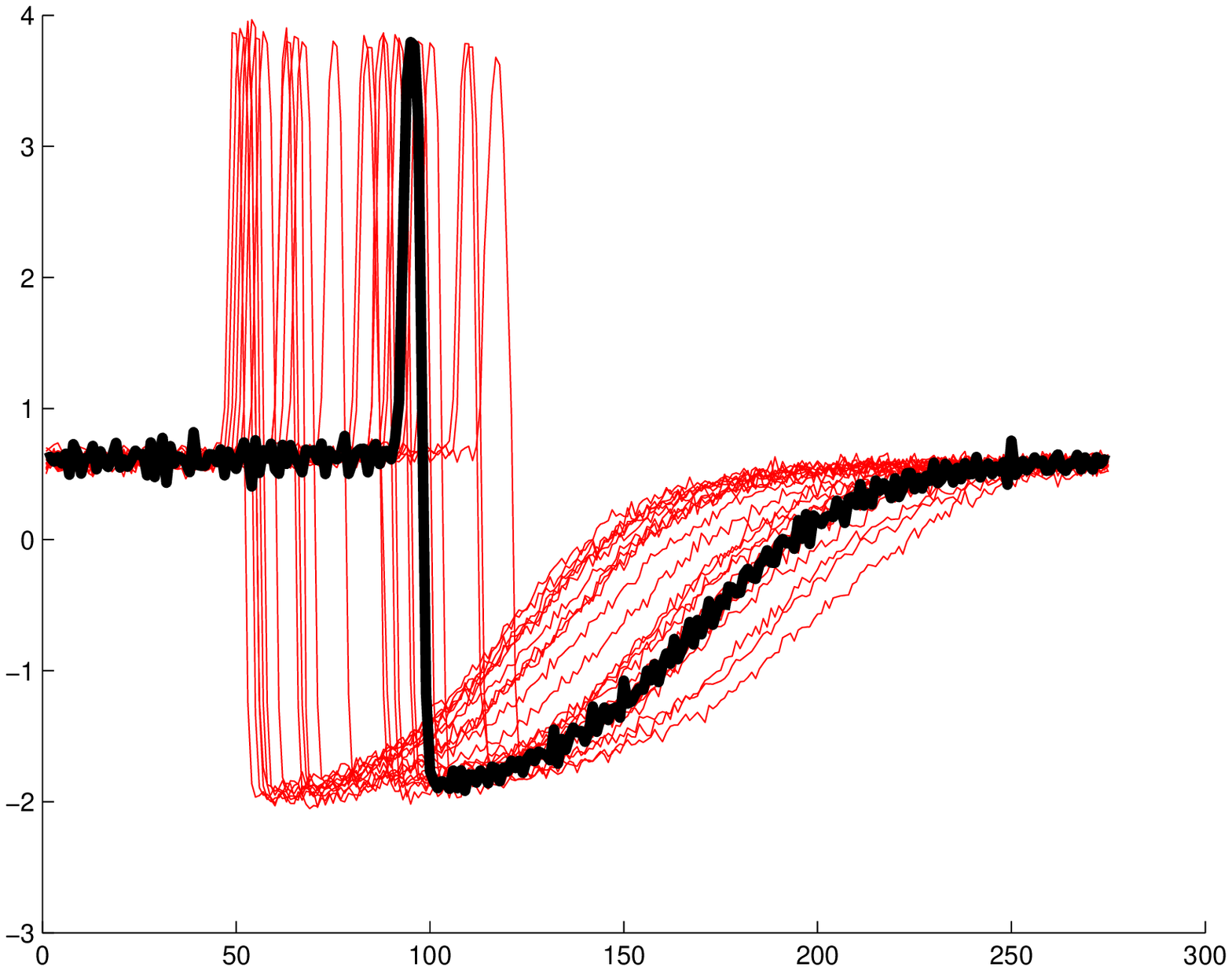}}\\
    {\includegraphics[scale=.25]{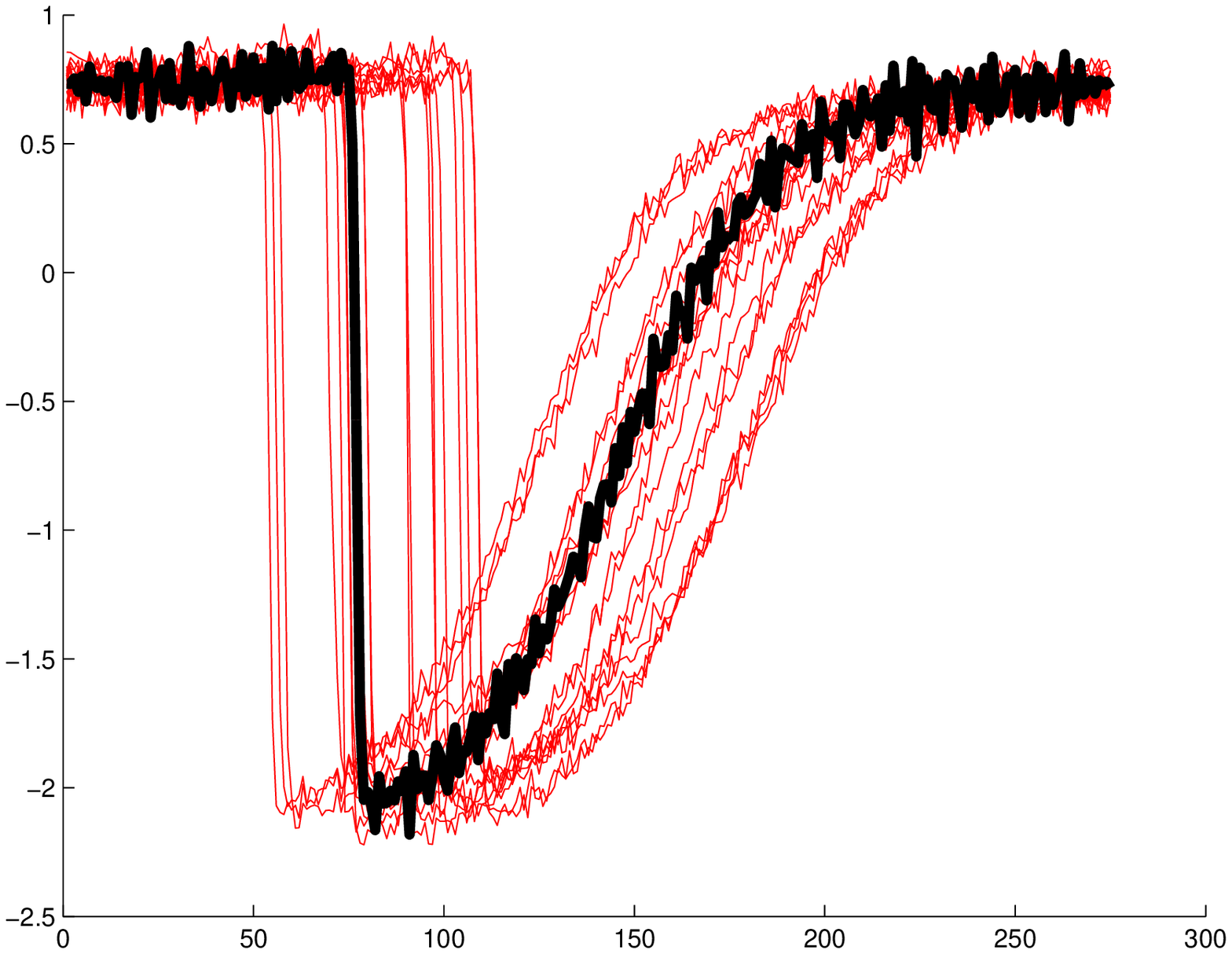}}
    	{\includegraphics[scale=.25]{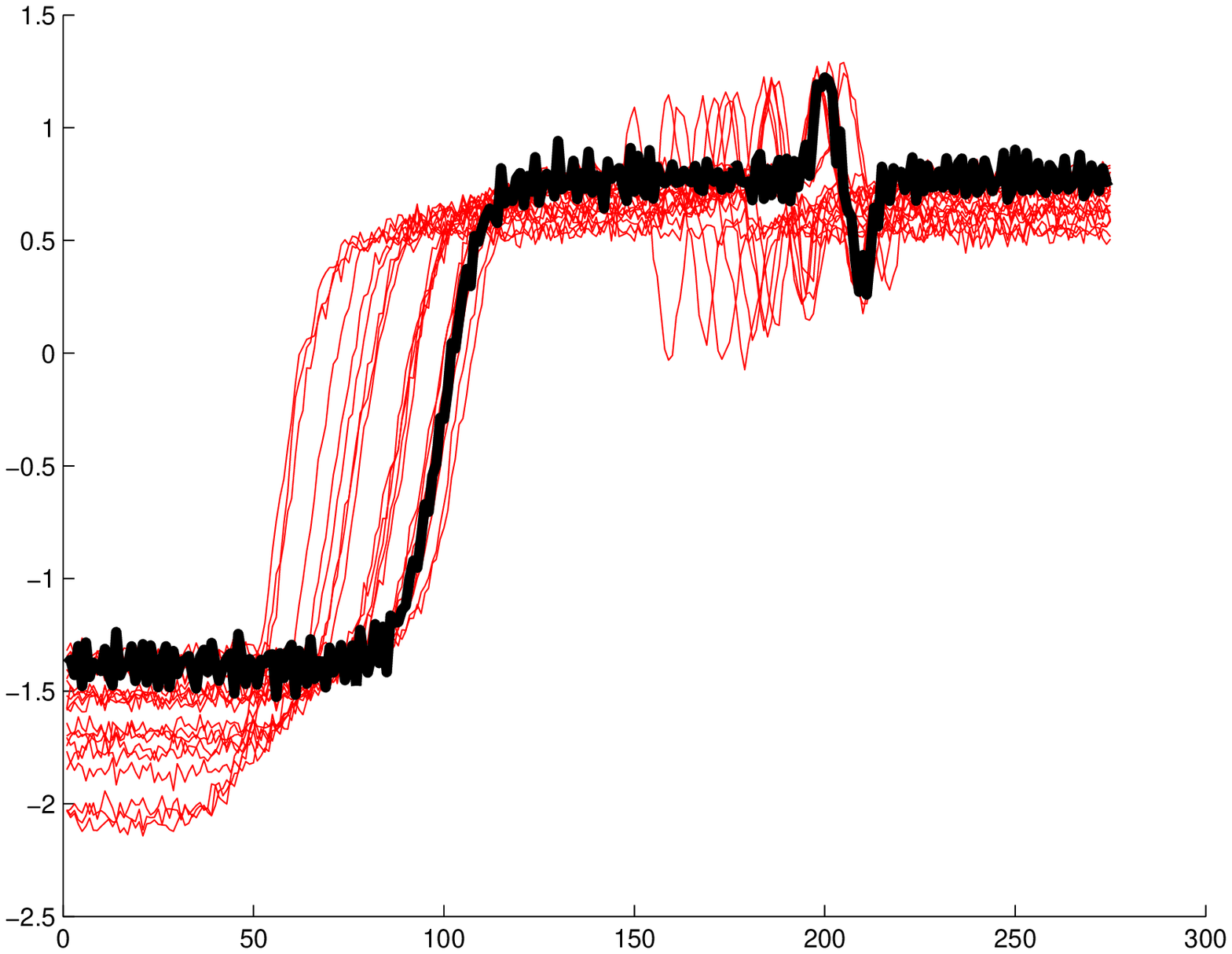}}
\end{tabular}
\caption{Centroid estimations of the first three categories (out of four) contained in the Trace dataset as a solution of the preimage problem. In bold blue, the centroid time series; in light red, the time series of the averaged dataset. Top left diagram of figure shows value of the minimized functional expressed on a log-scale, plotted against the iteration index.}

\label{fig:preimageTrace}
\end{figure}

For KDTW, the non-convex optimization problem cannot be straightforwardly addressed using gradient-based approaches mainly because the derivative cannot be determined analytically. Moreover, the number of variables (linear with the length of the time series and with the dimensionality of each sample) is generally high so this approach often encounters combinatorial difficulties related to the number of local minima. A derivative-free method could nevertheless be applied for local modelling of the functional to be optimized. In an attempt to carry out such a preimage formulation to estimate the time elastic centroid for a set of time series, we applied the state-of-the-art BOBYQA algorithm developed for bound constrained optimization without using derivatives \cite{Powel2009}. Fig.\ref{fig:preimageCBF} and Fig.\ref{fig:preimageTrace} give the centroid estimations for each category of the CBF and Trace datasets, respectively \cite{KeoghUCRdataset}. On the top left diagram of the figures, the values of the function to be minimized are plotted against the number of iterations. The optimization process is initialized using the medoid for each category. We show that the required number of iterations is quite high and depends on the number of variables. For the CBF dataset, the time series is made up of 128 samples while there are 275 samples for the Trace dataset. The convergence rate is roughly ten times slower for the Trace data set compared with the CBF dataset, mainly because KDTW complexity is quadratic with the length of the time series. The iteration cost becomes somewhat prohibitive for long time series or large time series datasets. Although this approach could be possibly optimized, the parameters need to be carefully set up (basically, definition of the trust region) and, in any case, as stated above, the optimum so provided remains an estimation of the centroid that is sought. Finally, note that the functional starts to decrease after attaining the number of iterations (in this case, twice the length of the time series) initially required for local estimation of the functional.

\section{Probabilistic interpretation of time elastic kernel alignment matrices}
In this section, we consider the recursive term $K^{xy}_{dtw}(.,.)$ that is used in Eq. \ref{Eq.KDTW}. When evaluating the similarity between two time series $X_{p}$ and $Y_{q}$ with respective lengths of $p$ and $q$, this recursion allows the construction of an alignment matrix $AM(i,j)$ with $i \in \{1 \cdots p\}$ and $i \in \{1 \cdots q\}$. The cell at location $(i,j)$ contains the summation of the global costs of all alignment paths, as defined in \textit{definition} \ref{def:alignmentPath}, that connect cell $(1,1)$ with cell $(i,j)$. For any alignment path $\pi$, the global cost is expressed as: 
\begin{equation}
cost(\pi)=\prod_{k=1}^{|\pi|} e^{-\nu d_{E}^{2}(X(\pi_k(1)),Y(\pi_k(2)))}
\end{equation}
i.e. the product along the path of the local alignment costs. We can give a probabilistic interpretation of these local costs $exp(-\nu d_{E}^{2}(X(\pi_k(1)),Y(\pi_k(2))))$: basically, we can assume that these local costs correspond (within the magnitude of the scalar multiplication constant) to the local \textit{a priori} probability of aligning sample $X(\pi_k(1))$ with sample $Y(\pi_k(2))$. By making this assumption, we eventually attach a probability distribution to the set of all alignment paths, with the $cost(\pi)$  corresponding (within the magnitude of the scalar multiplication constant) to the probability attached to alignment path $\pi$.

Hence, the cell $(i,j)$ of matrix $AM$, contains the sum of the probabilities (within the magnitude of the scalar multiplication constant) of the paths that connect cell $(1,1)$ to cell $(i,j)$.

Similarly, if, instead of $X$ and $Y$, we evaluate the similarity between $X_r$ and $Y_r$ derived from $X$ and $Y$ by reversing the temporal index, we obtain an alignment matrix $AM_r$ whose cell $(i,j)$ contains the sum of the probabilities (to within a multiplicative scalar constant) of the paths that connect cell $(p,q)$ tp cell $(i,j)$.

Finally, multiplying properly cells of $AM$ with cells of $AM_r$ yields the Alignment Matrix Average ($AMA$) defined as:
\begin{equation}
AMA(i,j)=AM(i,j) \cdot AM_r(p-i+1, q-j+1)
\label{eq:AMA}
\end{equation}
 and whose cell $(i,j)$ contains  the sum of the probabilities (upto the normalization constant) of the paths that connect cell $(1,1)$ to cell $(p,q)$ while going through the cell $(i,j)$.
 
From this path probability distribution, we can now derive an alignment probability distribution between the samples of $X$ and the samples of $Y$ as follows:

\begin{itemize}
\item For all $i$, the probability of aligning sample $X(i)$ is  $P(i)=1$; all samples need to be aligned.
\item Similarly, for all $j$, the probability of aligning sample $Y(j)$ is  $P(j)=1$.
\item The probability of aligning sample $X(i)$ with sample $Y(j)$ is $P(i,j)=P(i|j).P(j)=P(i|j)$. $P(i|j)$ is the probability that sample $X(i)$ is aligned with sample $Y(j)$ given that the alignment process is in state $j$. The estimation of $P(i|j)$ is obtained by using matrix $AMA$:
\[
P(i|j)= \frac{AMA(i,j)}{\sum_{i=1}^p AMA(i,j)}
\]
\item Furthermore, the probability of aligning sample $X(i)$  with sample $Y(j)$ is also $P(i,j)=P(j|i).P(i)=P(j|i)$. Similarly, the estimation of $P(j|i)$ is obtained by using matrix $AMA$:
\begin{equation}
\label{eq:Pij}
P(j|i)= \frac{AMA(i,j)}{\sum_{j=1}^q AMA(i,j)}
\end{equation}
\end{itemize}

Note that the normalization constant mentioned above is eliminated.

Since $P(i,j)=P(i|j)=P(j|i)$, we can finally estimate the probability of aligning sample $X(i)$ with sample $Y(j)$ as follows:
\begin{equation}
\label{eq:Pij}
P(i,j)= \frac{1}{2}\cdot\left(\frac{AMA(i,j)}{\sum_{i=1}^p AMA(i,j)}+ \frac{AMA(i,j)}{\sum_{j=1}^q AMA(i,j)}\right)
\end{equation}

Eq. \ref{eq:Pij} forms the basis of our pairwise time elastic time series averaging algorithm given below. 

As an example, Fig \ref{fig:SinTest} presents the AMA matrix corresponding to the alignment of a positive halfwave with a sinus wave. The three potential alignment pathes are clearly identified in the light blue and red colors.

\begin{figure}[ht!]
\centering
\begin{tabular}{ccc}
	\includegraphics[scale=.3, angle=0]{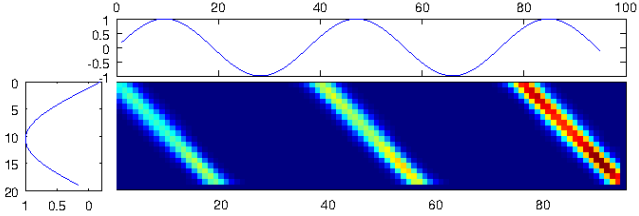} &
\end{tabular}
\caption{AMA matrix for the alignment of a positive halfwave with a sinus wave.}
\label{fig:SinTest}
\end{figure}

\section{Time elastic centroid based on the $AMA$ alignment matrix}
Based on the structure of the KDTW kernel and the AMA matrix, and by using the so-called DtwBarycenter Averaging (DBA) method developed by \cite{Abdulla2003}, \cite{Petitjean2011}, \cite{Hautamaki2008}, we first present the  KernelDtwNarycenter Averaging (KDBA) algorithm for estimating a time elastic centroid for a set of time series according to an iterative agglomerative procedure as shown in Fig. \ref{fig:hac-iter-b}. Secondly, we detail the concept of a time elastic average for a pair of time series (KDTW-PWA),  and then develop the progressive heuristic approach presented in Fig. \ref{fig:hac-iter-a} that uses KDTW-PWA to estimate another kind of time elastic centroid (KDTW-C1) for a set of time series of any cardinal. 

\subsection{KDTW-Centroid of a set of time series based on KDBA algorithm}
Following the DBA algorithmic approach \cite{Abdulla2003},\cite{Hautamaki2008}, we present here the development of our kernelized version called KDBA. KDBA directly applies the definition of the alignment matrix average (AMA) as given in Eq.\ref{eq:AMA} and its probabilistic interpretation Eq.\ref{eq:Pij}.

    \begin{algorithm}[h!]
      \caption{KDBA}\label{alg:KDBA}
      \begin{algorithmic}[1]
        \Procedure{KDBA}{$R$,$S$, $\nu$}
        \State // R: a reference time series
        \State // S: a set of time series $\{S_1, \cdots, S_N\}$
        \State // $\nu$: the stiffness parameter of KDTW kernel 
        \State Double AMA(.,.);
        \State Vector-Of-SetOfSamples SampleAssociations(L); 
        \State Ts $A(|R|)$; //Create a D dimensional  
        \State \hspace{14mm}//time series of length $L$;
        \For{Int $i=1$ to $|R|$}
          SampleAssociations(i)=\{\};
        \EndFor
        \For{Int $n=1$ to $|S|$}
          \State  Evaluate $AMA$ matrix for $R$, $S_n$ with $\nu$;
          \State Ts $ts$//containing L "zeroed" samples;
          \State Double $normFactor(|R|)$; 
          \For{Int $i=1$ to $|R|$}
          	  \State normFactor(i)=0;
          	  \For{Int $j=1$ to $|S_n|$}
          	    \State $ts(i) = ts(i) + S_n(j)*AMA(i,j)$;
          	    \State $normFactor(i) = normFactor(i)+$
          	    \State  \hspace{14mm}$AMA(i,j)$;
          	  \EndFor
          	  \State $ts(i)=ts_1(i)/normFactor(i)$;
          	  \State SampleAssociations(i)=($ts(i)$);
          \EndFor
        \EndFor
        \For{Int $i=1$ to $|R|$}
    			\State A(i)=barycenter(SampleAssociations(i));
    		\EndFor
        \State \Return $A$
        \EndProcedure
      \end{algorithmic}
    \end{algorithm}

    \begin{algorithm}[h!]
      \caption{iKDBA}\label{alg:iKDBA}
      \begin{algorithmic}[1]
        \Procedure{iKDBA}{$C$,$S$, $\nu$}
        \State //C: a reference time series
        \State //S: a set of time series
        \State //maxIter: maximum number of iterations
        \State //$\nu$: the stiffness parameter of KDTW kernel 
        \State Ts $A$; //a D dimensional Timeseries 
        \State Double inertia = computeInertia$(C, S$);
        \State Boolean Continue=True;
        \State Int $i =0$;
        \While{Continue} 
            \State A=C;
            \State C=KDTW-C2($C$,$S$, $\nu$);
            \State Double new\_inertia = computeInertia$(C, S)$;
            \If{new\_inertia $>$ inertia OR $i>maxIter$}
            		\State Continue = False;
            \EndIf
            \State i=i+1;
        \EndWhile
        \State \Return $A$
        \EndProcedure
      \end{algorithmic}
    \end{algorithm}

Let us consider a set $S$ of $N$ time series, $S=\{S_1, S_2, \cdots, S_N\}$, and $R$ a reference time series. Let $|R|$ and $|S_n|$ be the lengths of $R$ and $S_n$, respectively. $P_n(i,j)$, with $i=1\{1, ...|S_n|\}$ and $j=1\{1, ...|R|\}$, is obtained from the AMA matrix resulting from the alignment of $S_n$ with $R$, according to Eq.\ref{eq:Pij}.  Algorithm \ref{alg:KDBA} computes an average time series $A$ according to the following equation:
\begin{equation}
\forall i \in \{1, \cdots, |r|\},\hspace{1mm} \texttt{A}(i)=\frac{1}{N} \sum_{n=1}^N\sum_{j=1}^{|S_n|}P_n(i,j)S_n(j)
\end{equation}

Note that the iterative average of time series produced by algorithm \ref{alg:KDBA} has the same size as the reference time series $R$.

The algorithm \ref{alg:KDBA} can be refined by iterating until no further improvement is obtained \cite{Petitjean2011}. An improvement is observed when the sum of the distances (resp. similarities) between the current average $R$ and the new pairwise average provided by KDBA, $A$, is lowered (resp. increased). Algorithm \ref{alg:iKDBA} implements this iterative strategy, which will necessarily find a local minimum or will stop when a maximum number of iterations has been reached.

    \begin{figure}[h!]
\centering
    	\includegraphics[scale=.4]{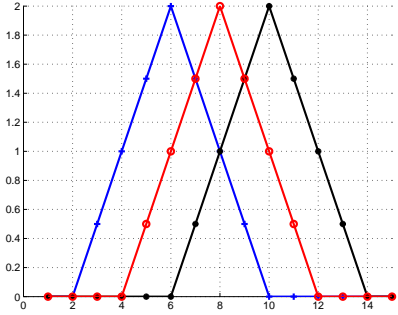} 
    	\label{fig:time-shift}
    	\caption{Expected time location for the centroid (in red circles) of two triangular-shaped time series shifted in time (in blue '+' and black '*')}
\end{figure}   
\subsection{KDTW average of a pair of time series (KDTW-PWA)}

    \begin{algorithm}[H]
      \caption{KDTW-PWA}\label{alg:KDTW-PWA}
      \begin{algorithmic}[1]
        \Procedure{KDTW-PWA}{$X$,$Y$, $AMA$}
        \State //X,Y: two time series of $D$ dimensional samples
        \State //AMA: the average alignment matrix for X,Y
        \State Int $p=|X|$, $q=|Y|$, $L=\text{max}\{p,q\}$;
        \State Ts $A(L)$, $B(L)$; //Create 2 D dimensional  
        \State \hspace{27mm}//time series of length $L$;
        \State Double $\alpha$;
        \State Double $N_A(L), N_B(L)$; //two double arrays
        \For{Int $i=1$ to $L$}
        		\For{d=1 to D}
        			\State $A(i,d)=0$, $B(i,d)=0$;
        		\EndFor
        		\State $N_A(i)=0$, $N_B(i)=0$;
        	\EndFor	
        \For{Int $i=1$ to $L$}    		
        		\If{$i<p$}
        		  \For{Int $j=1$ to $q$}
        		  	\State $\alpha=(i+j)/2 - \lfloor(i+j)/2\rfloor$;
        		  	\For{d=1 to D}
        		  		\State $A(\lfloor(i+j)/2\rfloor,d) +=$
        		  		\State \hspace{0mm}$\alpha\cdot(X(i,d)+Y(j,d))\cdot AMA(i,j)$;
        		  		\State $A(\lceil(i+j)/2\rceil,d) +=$
        		  		\State \hspace{0mm}$(1-\alpha)\cdot(X(i,d)+Y(j,d))\cdot AMA(i,j)$;
        		  	\EndFor
        			\State  $N_A(\lfloor(i+j)/2\rfloor)+=\alpha*AMA(i,j)$;
        			\State  $N_A(\lceil(i+j)/2\rceil)+=(1-\alpha)*AMA(i,j)$;
        		  \EndFor
        		\EndIf
        	    \If{$i<q$}
        		  \For{Int $j=1$ to $p$}
        		  	\State $\alpha=(i+j)/2 - \lfloor(i+j)/2\rfloor$;
        		  	\For{d=1 to D}
        		  		\State $B(\lfloor(i+j)/2\rfloor,d) +=$
        		  		\State \hspace{0mm}$\alpha\cdot(X(j,d)+Y(i,d))\cdot AMA(j,i)$;
        		  		\State $B(\lceil(i+j)/2\rceil,d) +=$
        		  		\State \hspace{0mm}$(1-\alpha)\cdot(X(j,d)+Y(i,d))\cdot AMA(j,i)$;
        		  	\EndFor
        			\State  $N_B(\lfloor(i+j)/2\rfloor)+=\alpha*AMA(j,i)$;
        			\State  $N_B(\lceil(i+j)/2\rceil)+=(1-\alpha)*AMA(j,i)$;
        		  \EndFor
        		\EndIf
        \EndFor
        \For{Int $i=1$ to $L$}
        		\For{d=1 to D}
        			\State $A(i,d)=(A(i,d)/N_A(i)+B(i,d)/N_B(i))/4$;
        		\EndFor
        	\EndFor
        \State \Return $A$
        \EndProcedure
      \end{algorithmic}
    \end{algorithm}

Similarly to DBA, the KDBA algorithm averages a set of time series in the sample space but not along the time axis. Basically, let us suppose we are averaging two triangular-shaped time series such as represented  by the blue crosses and black dots on Fig.\ref{fig:time-shift}. When using DBA or KDBA algorithms with one of the two time series acting as the reference, then the calculated average would be the reference distribution itself. However, we would also expect to average the time shift between the two series, thus obtaining the distribution indicated by the red dots in Fig.{fig:time-shift}. This is precisely our main motivation for the deriving the following Pair Wise Averaging (KDTW-PWA) algorithm designed to average a pair of time series in the sample space but also along the time axis.

Algorithm \ref{alg:KDTW-PWA} provides the KDTW-PWA average ($A$) of the two time series $X$ and $Y$ according to Eq.\ref{eq:averageExpectation}.

\begin{eqnarray}
\label{eq:averageExpectation}
   \begin{array}{ll}
   
   \forall k=1 \cdots L \texttt{, } \displaystyle{A(k)=\sum_{i,j|\frac{i+j}{2}=k} \left(P(i,j)\cdot\frac{X(i)+Y(j)}{2}\right)}\\
   \hspace{6mm}\displaystyle{=\sum_{i,j|\frac{i+j}{2}=k} \left(\frac{P(i|j)+P(j|i)}{2}\cdot\frac{X(i)+Y(j)}{2}\right)}
\end{array}
\end{eqnarray}

\begin{figure}[h!]
\centering
\begin{tabular}{ccc}
    	\includegraphics[scale=.25]{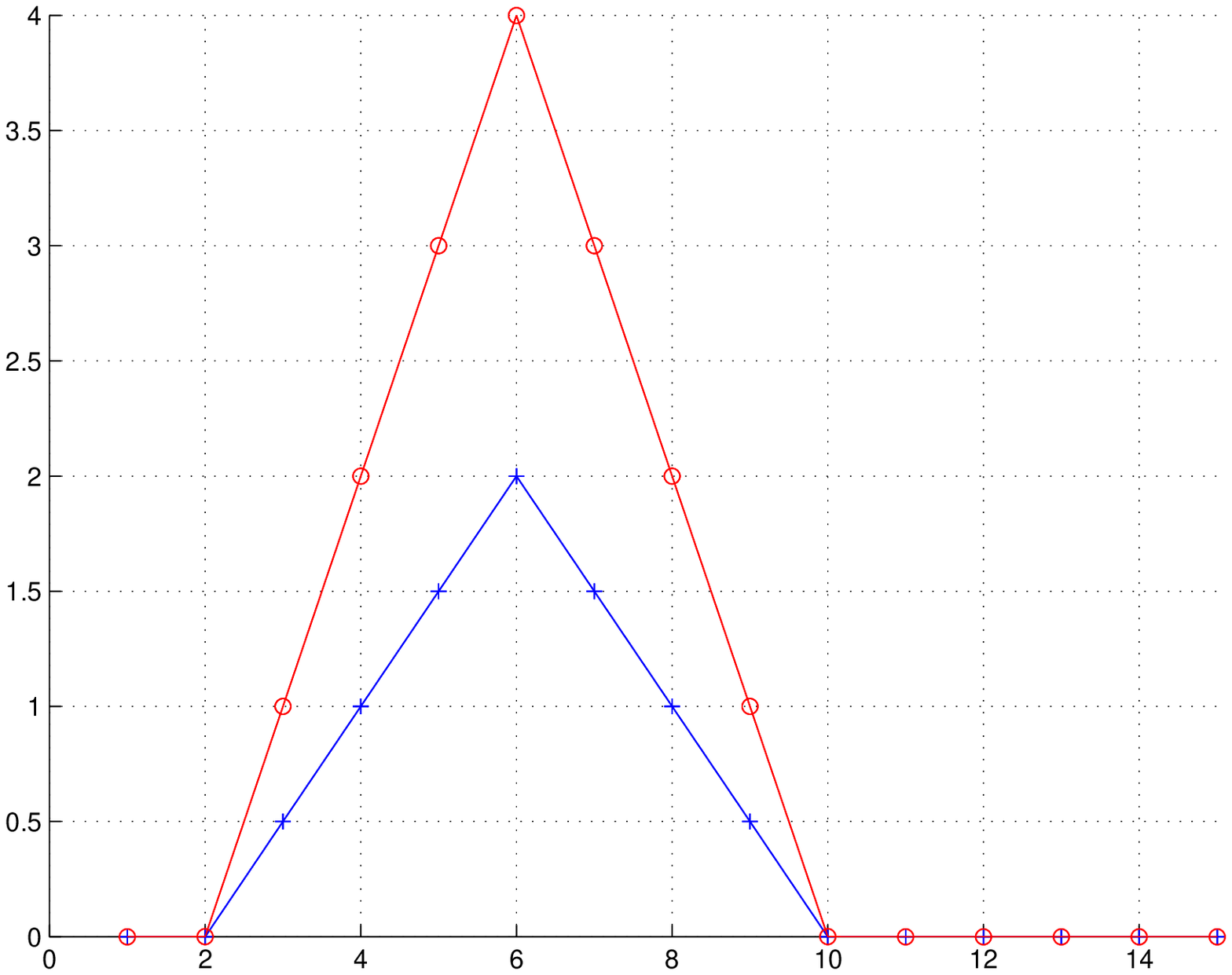} 
    	\includegraphics[scale=.25]{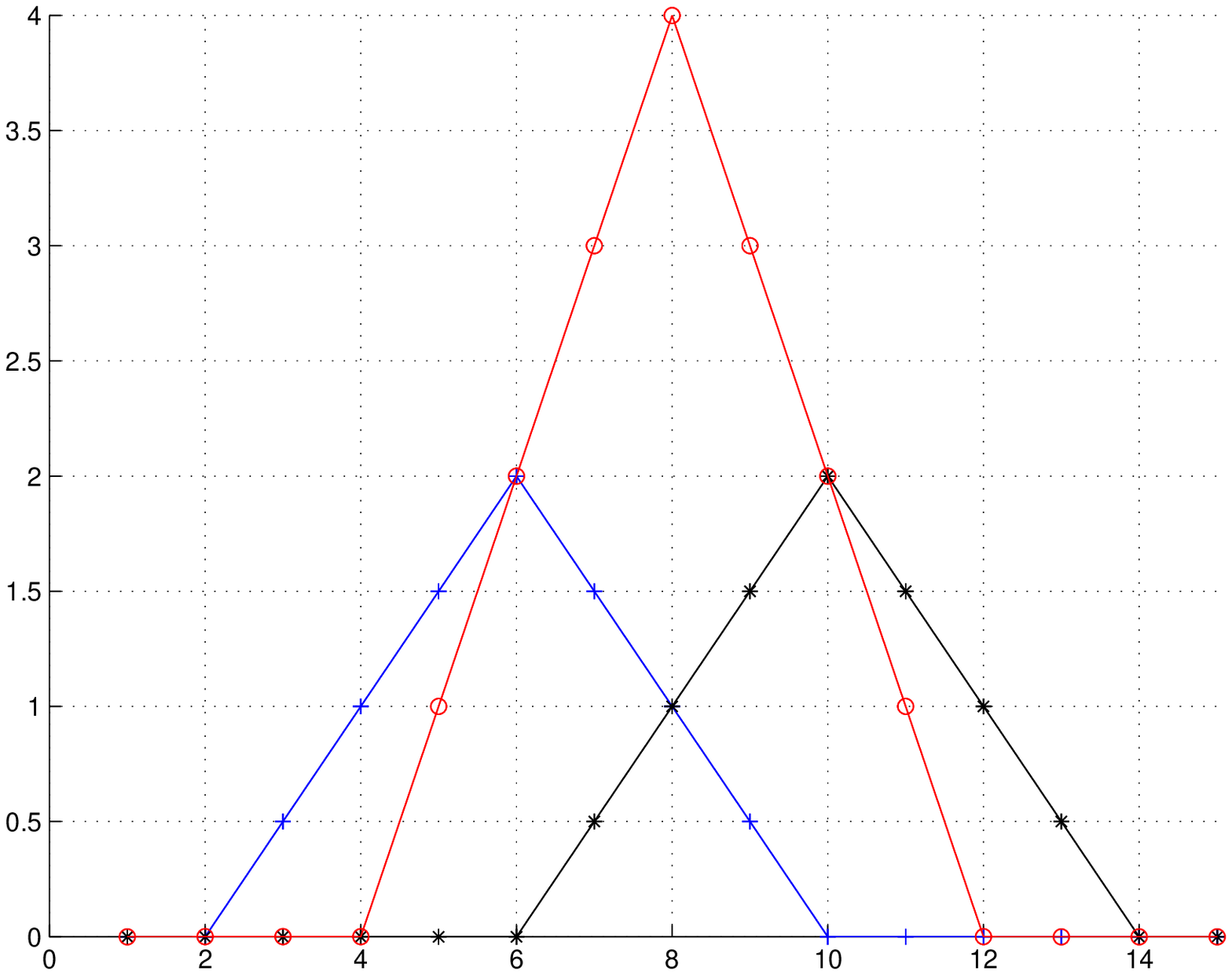}\\
    	\includegraphics[scale=.25]{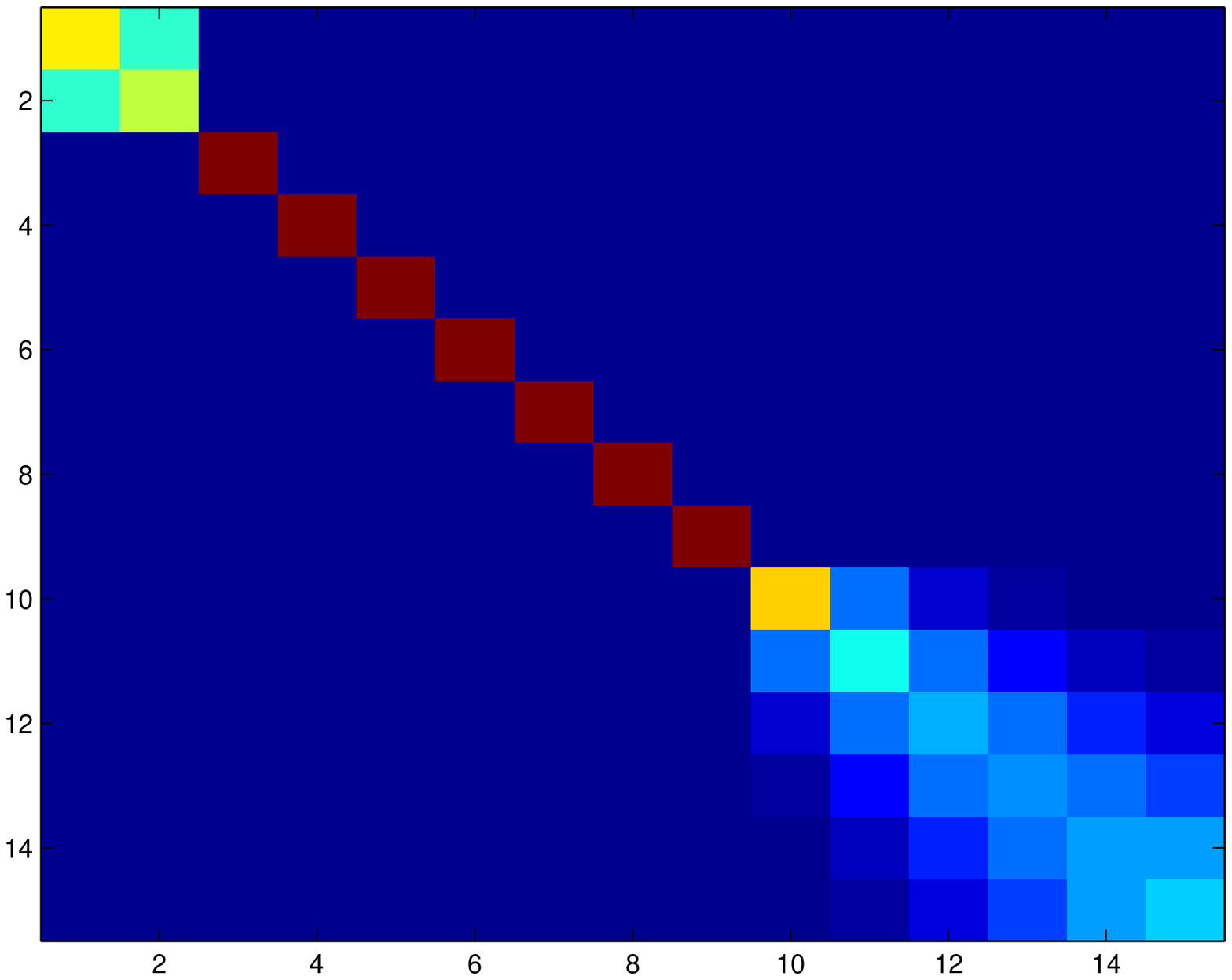}  
    	\includegraphics[scale=.25]{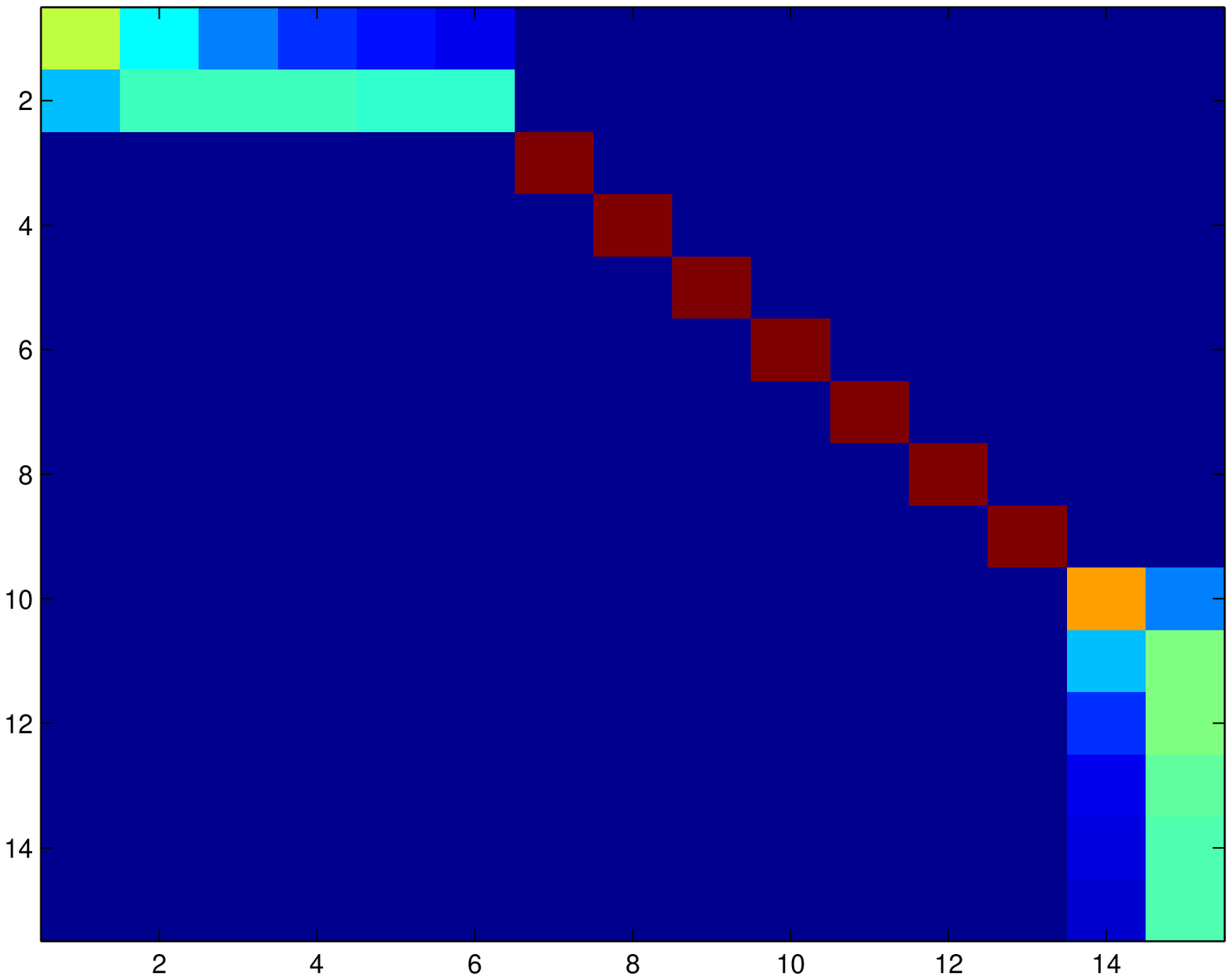} 
\end{tabular}
\caption{Averaging triangular-shaped time series. On the left, the two time series (in blue) are identical (superimposed) and the centroid (red) is amplified by a factor of two. On the right, the two time series (in blue) have the same shape but have been shifted in time. The KDTW-PWA average is given in red, still amplified by a factor of two.  The corresponding (normalized) AMA alignment matrices are given at the bottom. }
\label{fig:toyAverage}
\end{figure}

As the time indices are considered discrete (integer values), the time averaging $(i+j)/2$ is smoothed between the floor and cell integer values, using the smoothing coefficient $\alpha$ (line 17 of the algorithm).

Thus, the KDTW-PWA jointly averages the sample values of the two time series and their time locations. Eq. \ref{eq:averageExpectation} allows us to interpret the centroid of a pair of time series as the mathematical expectation of aligning the two sequences of samples.

\begin{figure}[ht!]
\centering
\begin{tabular}{ccc}
	\includegraphics[width=80mm, height=20mm, angle=0]{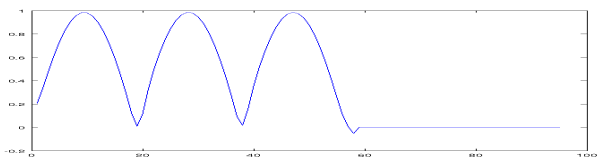} &
\end{tabular}
\caption{Centroid corresponding to the pairwise alignment for the sinus experiment depicted in Fig \ref{fig:SinTest}.}
\label{fig:SinCent}
\end{figure} 
As an example, the centroid corresponding to the pairwise alignment of the sinus experiment depicted in Fig \ref{fig:SinTest} is presented in Fig \ref{fig:SinCent}. Notice that in the centroid, the negative halfwaves of the sine wave have been \textit{filtered}. This is because the negative halfwaves do not match with the positive halfwave that is aligned with the sine wave. 

In Fig \ref{fig:toyAverage}, we present a very simple experiment that consists of  averaging two identical triangular-shaped time series (on left of figure) and two time series with identical triangular shapes but shifted in time. At the bottom of the figure, the corresponding $AMA$ matrices are presented. The KDTW-PWA distributions, presented in red, are multiplied by a factor of two to facilitate reading of the figure. We can see that, for both situations, the centroid is precisely located at the correct averaged time of occurrence of the two time series, whether or not they are shifted in time. The most likely alignment areas on the AMA matrices are shown in in red and the less likely alignment areas in blue. The time shift is clearly visible on the right-hand figure.

\begin{table*}[htb!]
\centering
\begin{tabular}{|c|c|c|}
\hline   
    \textbf{DBA} & \textbf{iKDBA} & \textbf{pKDTW-PWA} \\
\hline\hline 
    	\includegraphics[scale=.3]{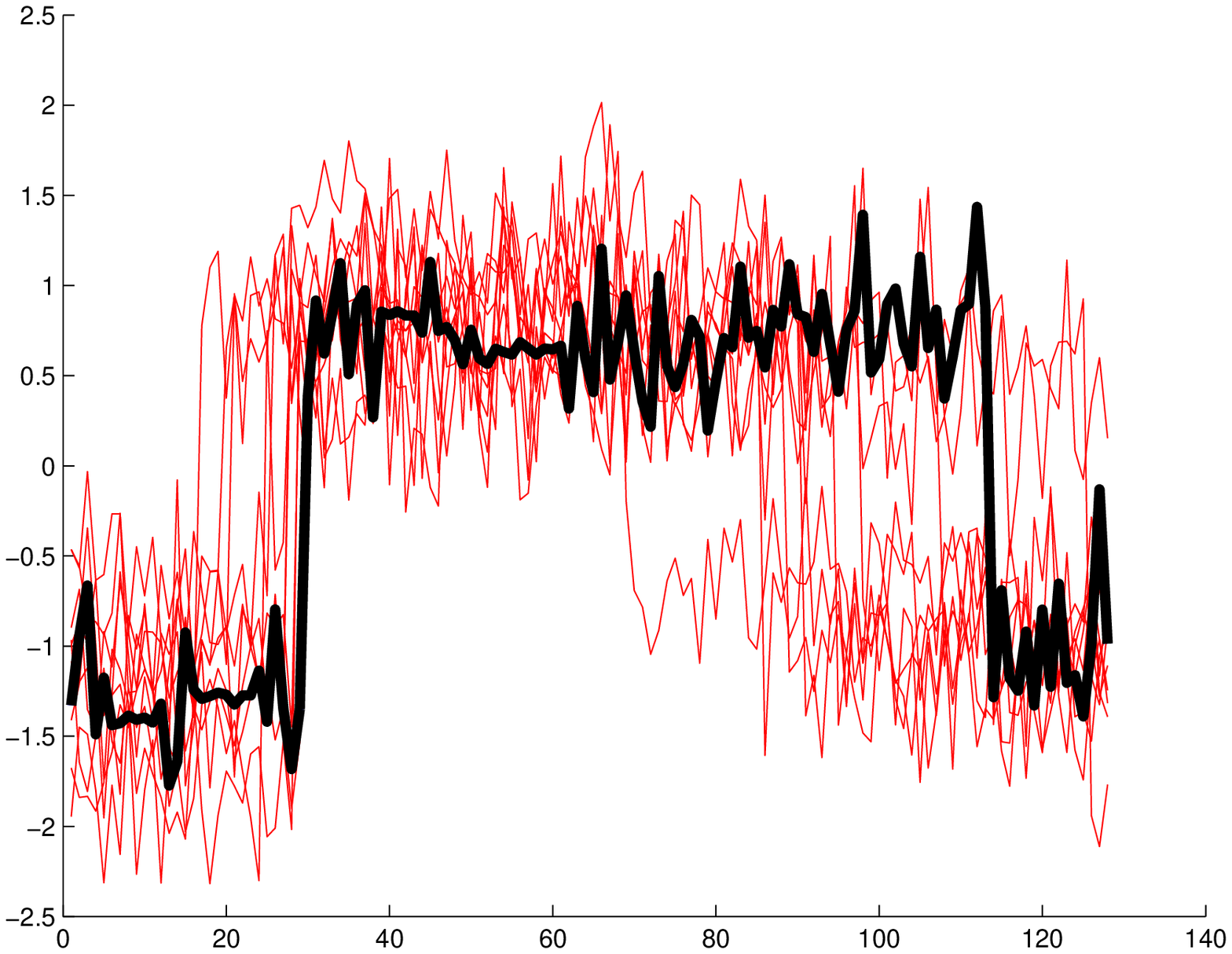}&
    	\includegraphics[scale=.3]{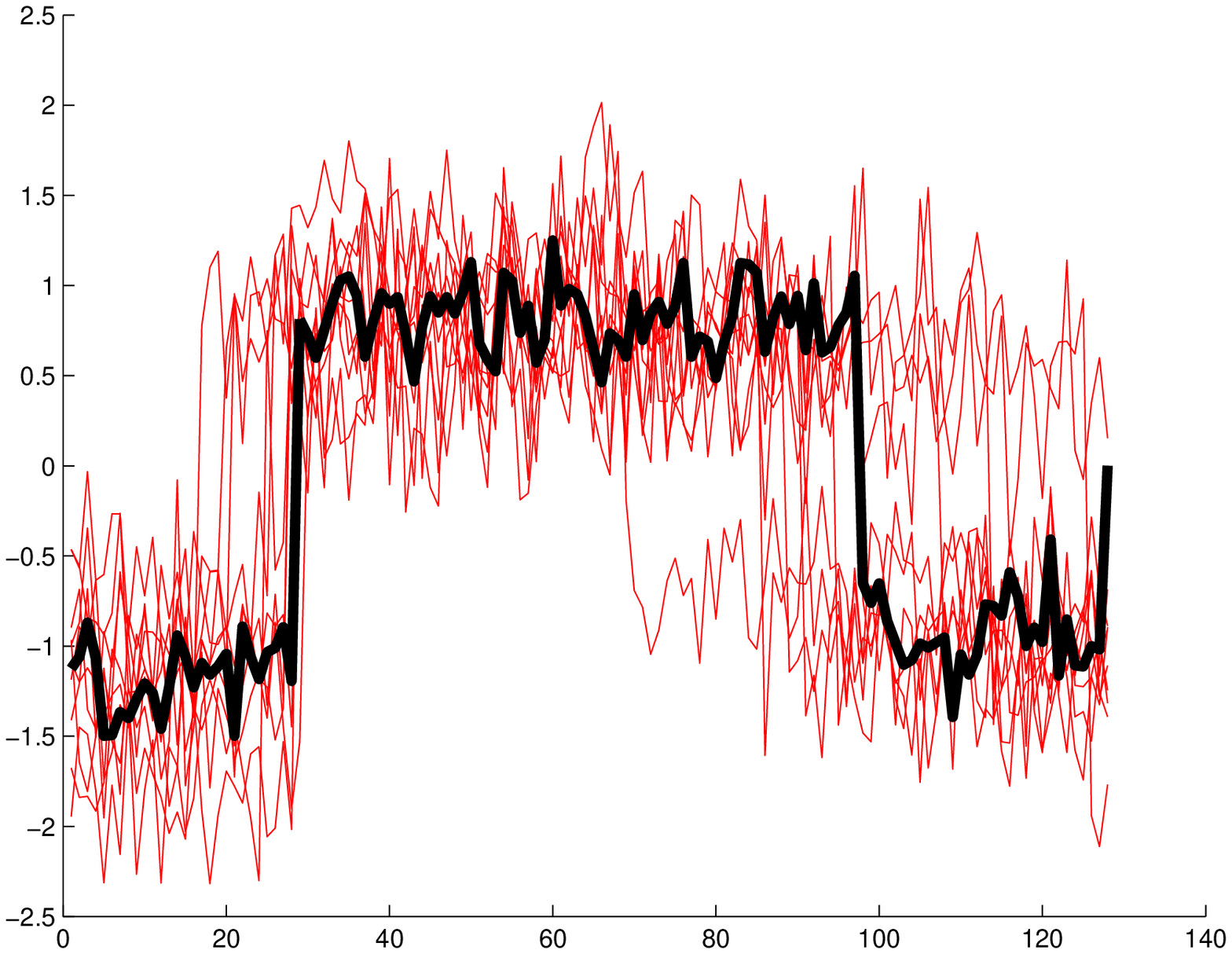}&
    	\includegraphics[scale=.3]{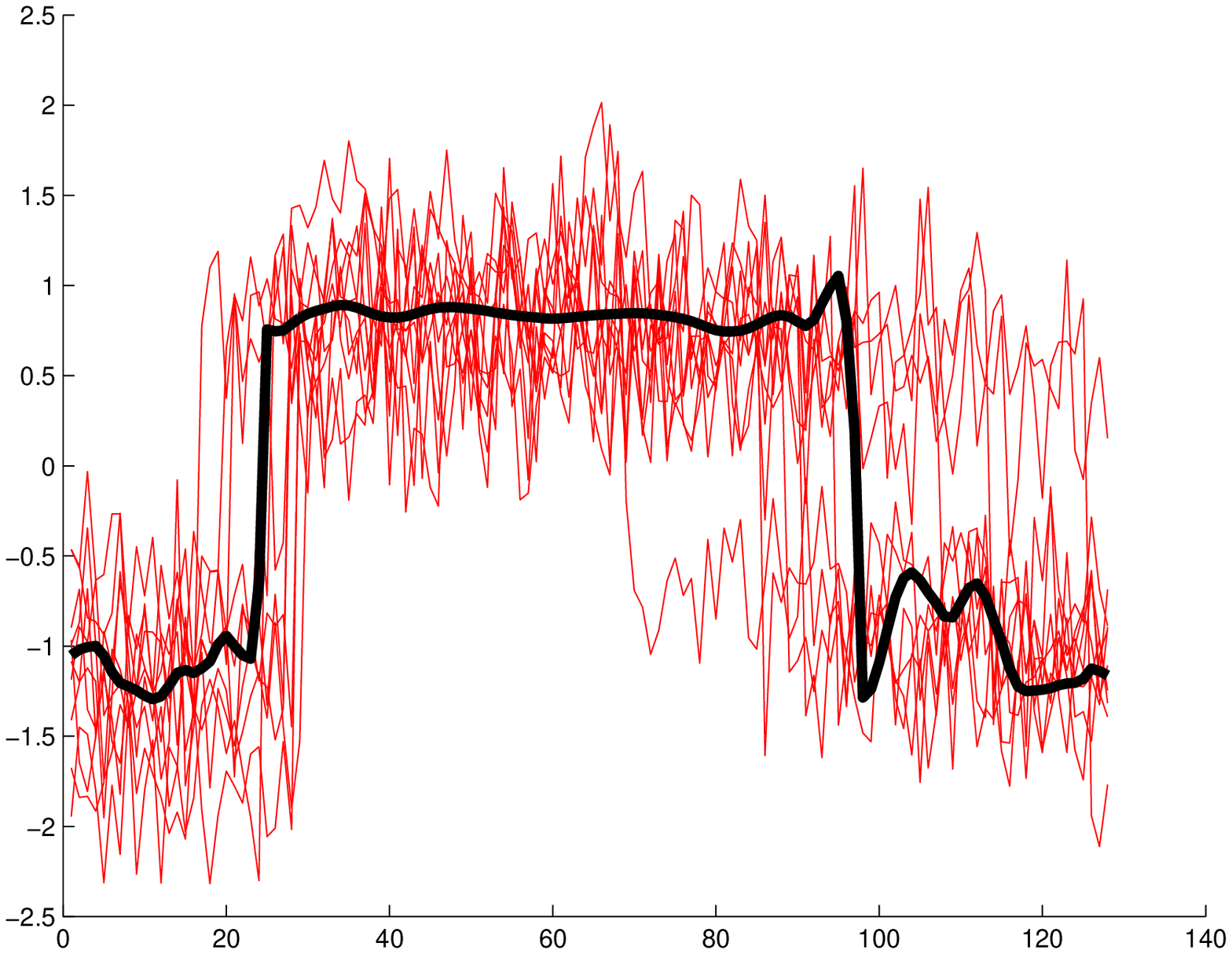}\\
    \includegraphics[scale=.3]{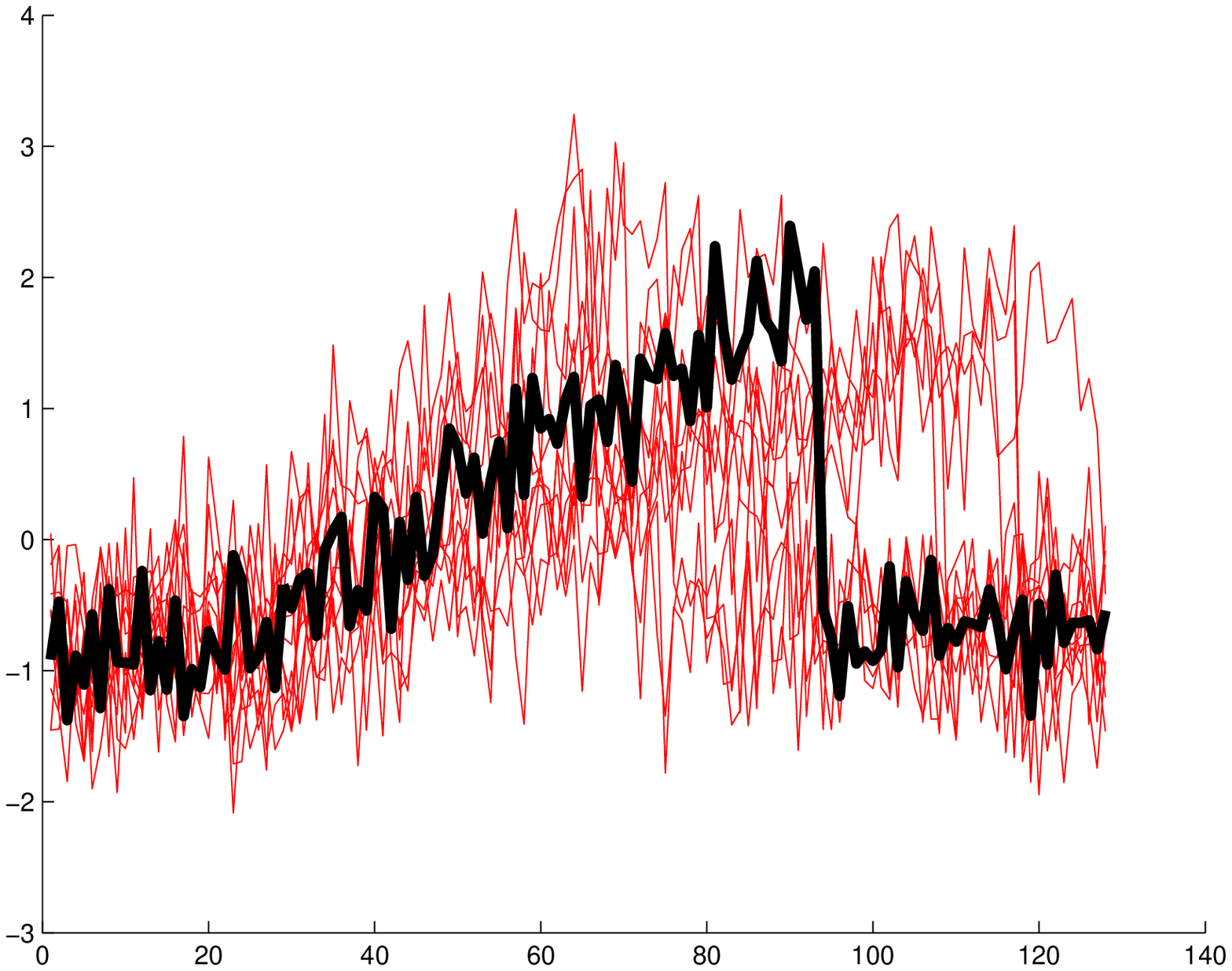}&
    \includegraphics[scale=.3]{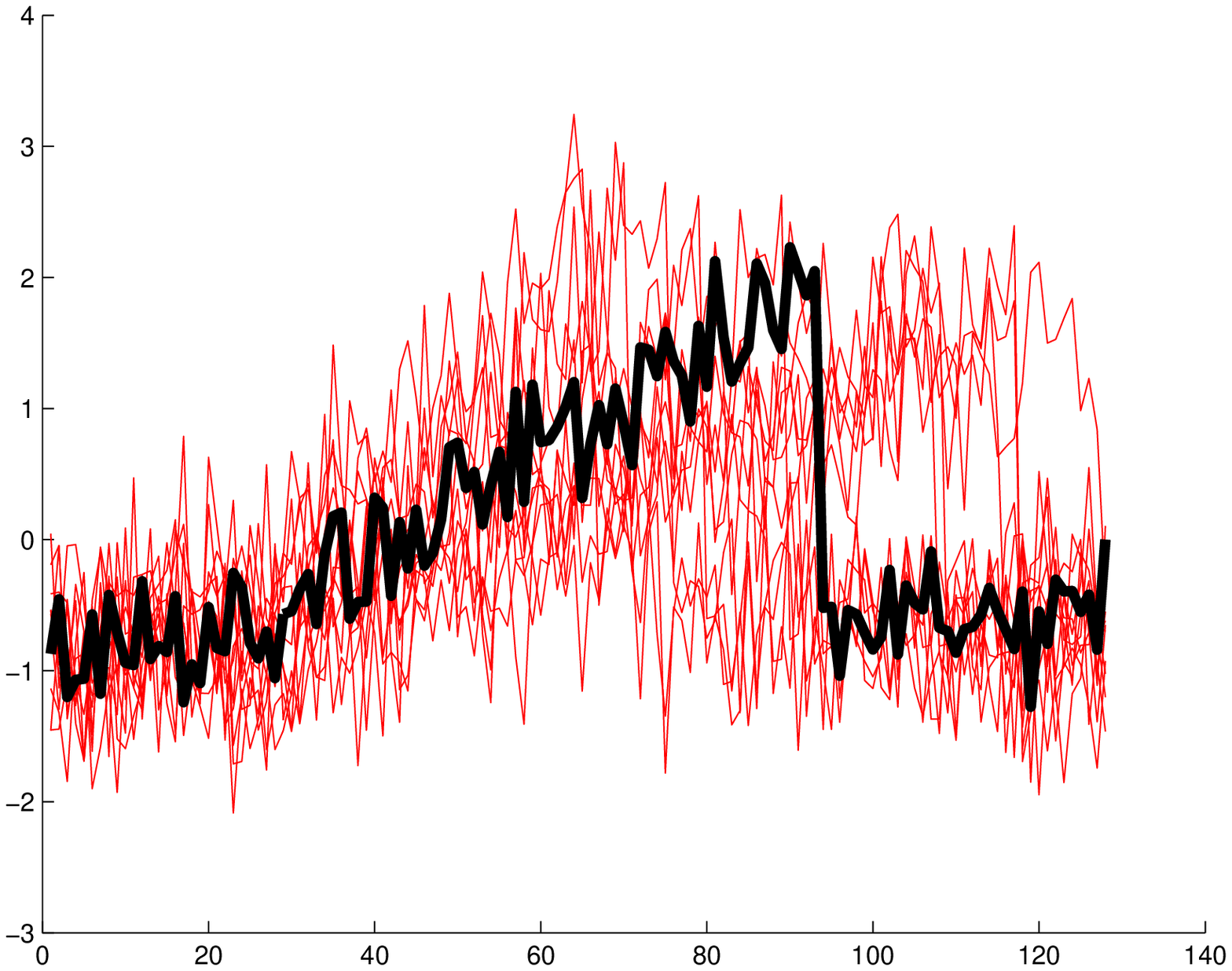}&
    \includegraphics[scale=.3]{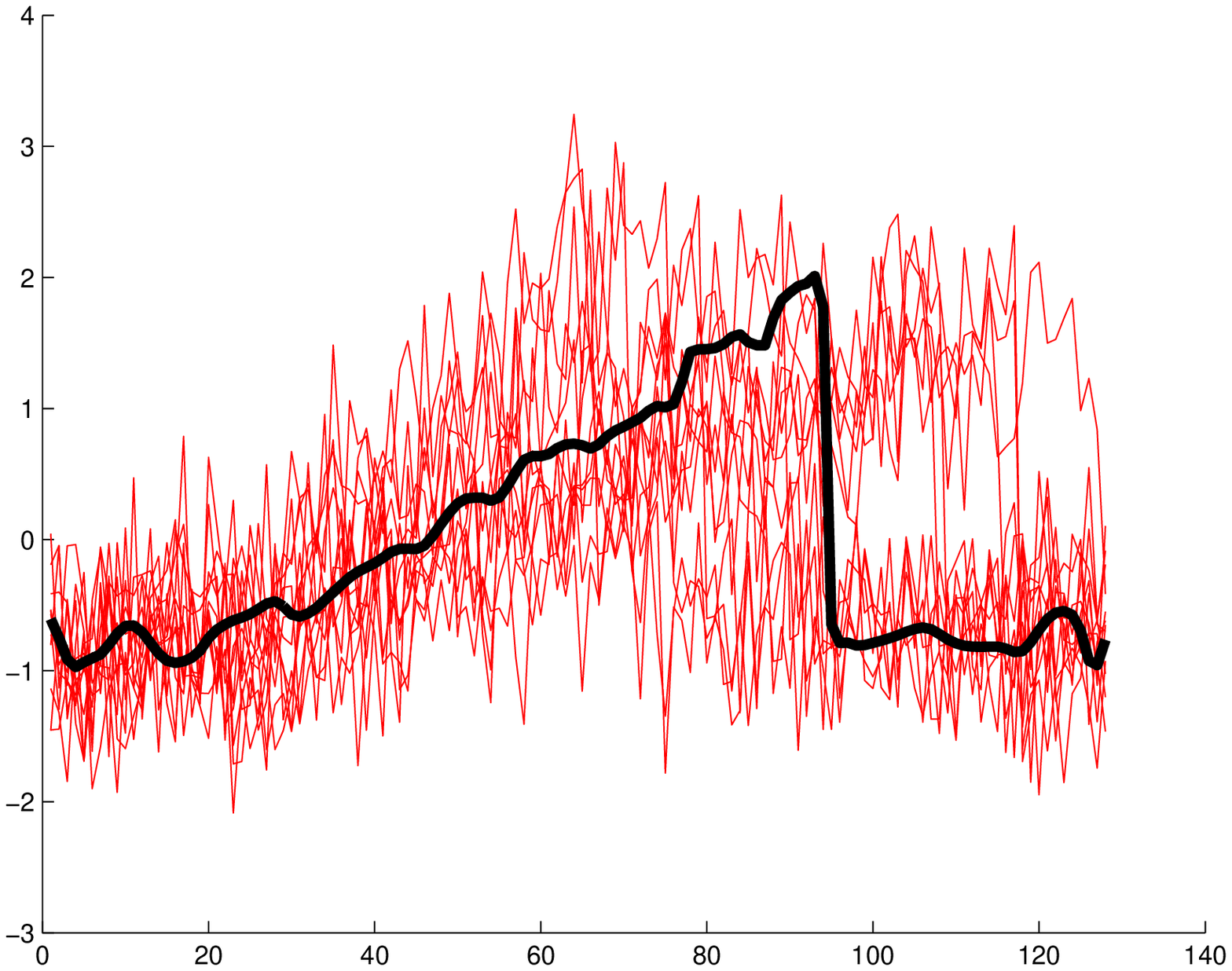}\\
    	\includegraphics[scale=.3]{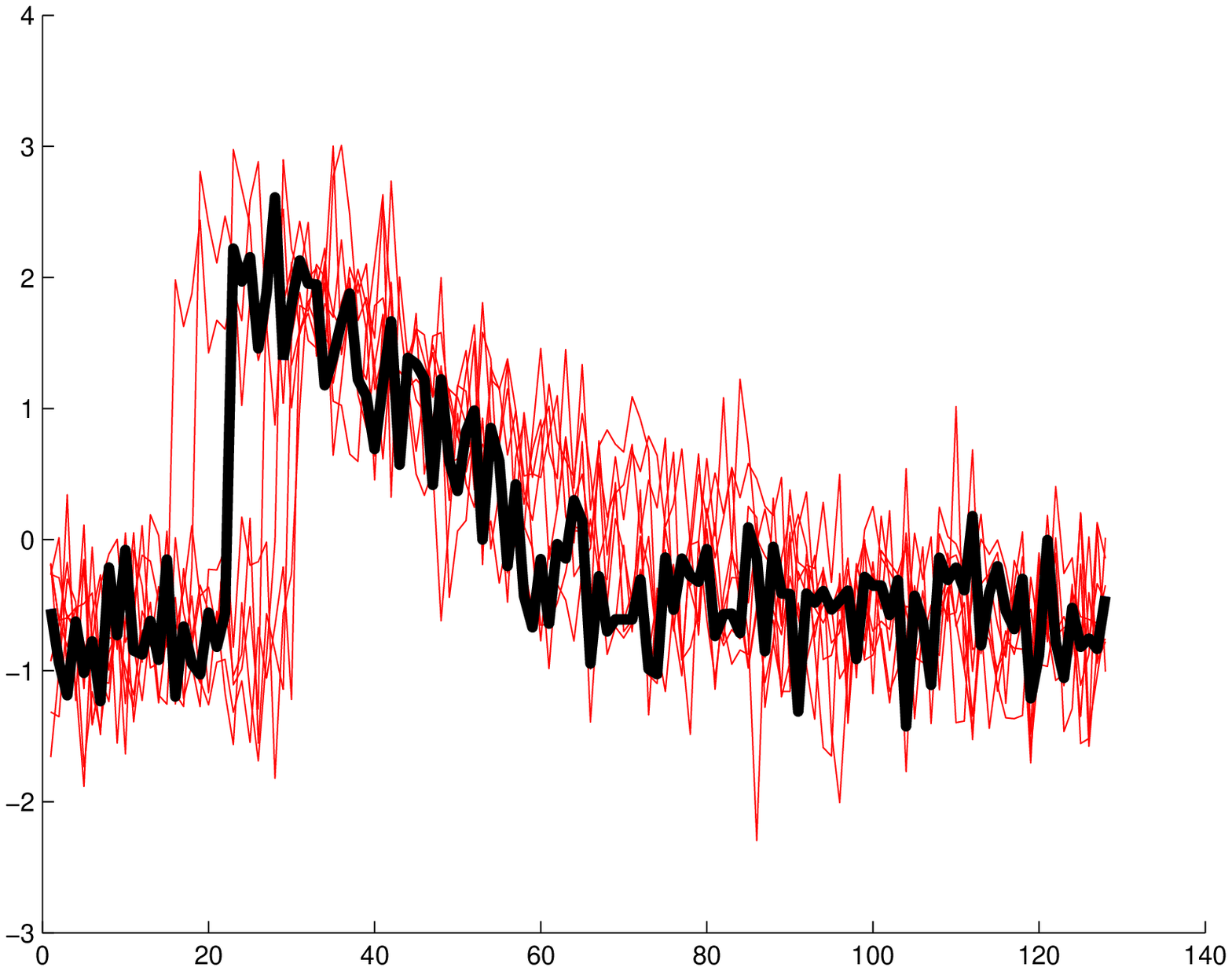}&
    	\includegraphics[scale=.3]{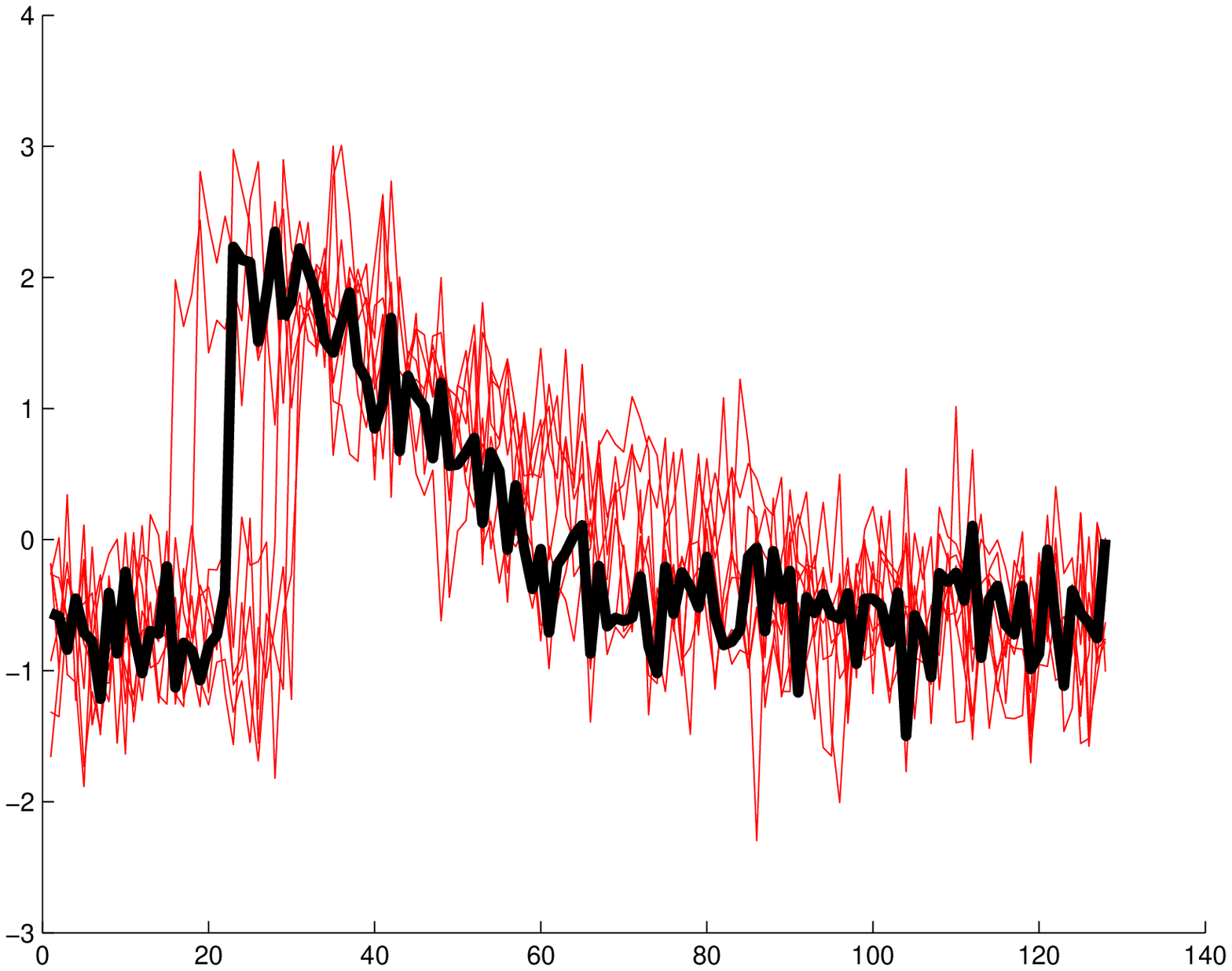}&
    	\includegraphics[scale=.3]{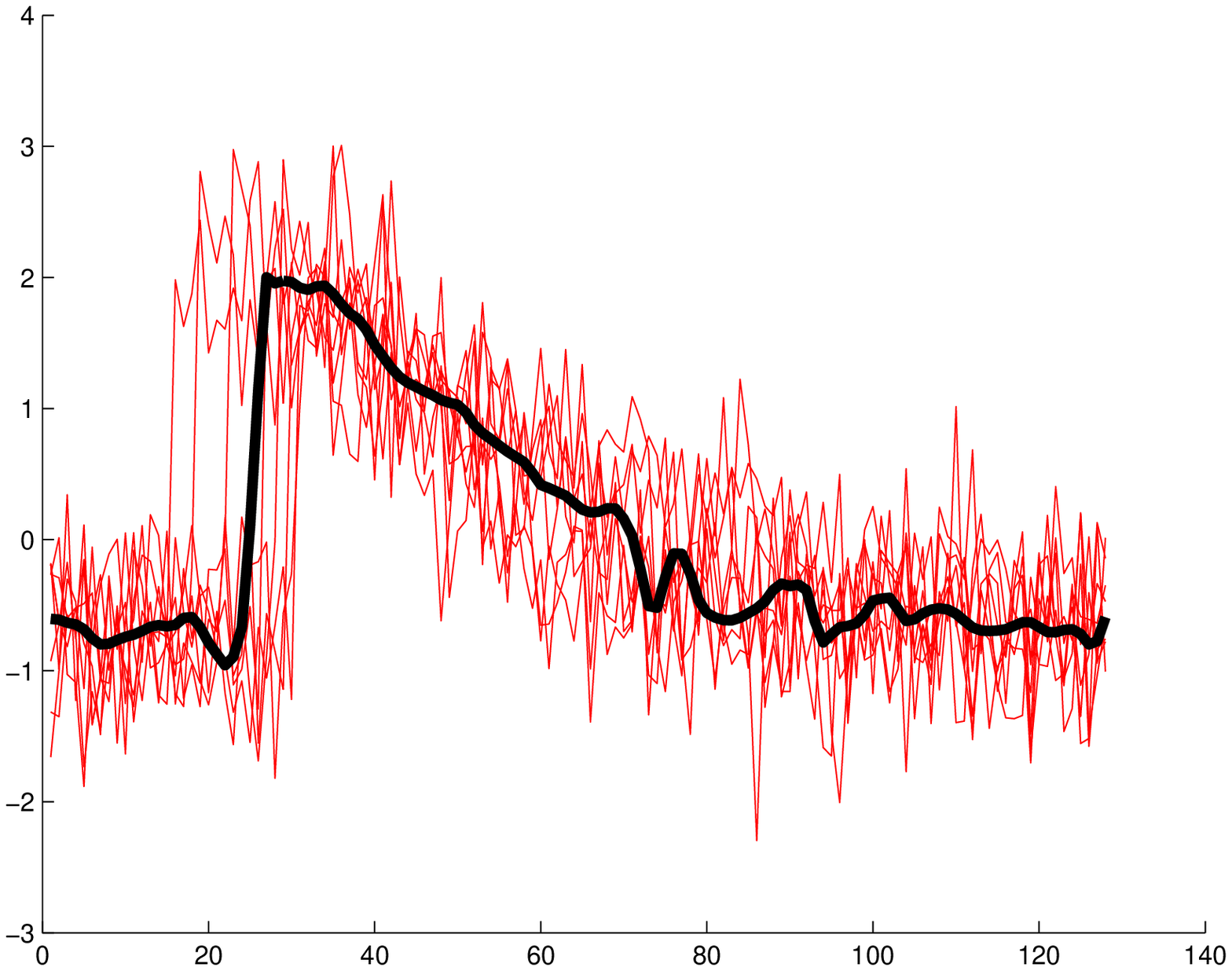}\\
\end{tabular}
\caption{Centroid estimation for the three categories of the CBF dataset. The centroid estimation is indicated as a bold black line superimposed on of the time series (in light red) that are averaged. The centroid estimates provided by the DBA algorithm are given on the left side, the estimates provided by the iKDBA algorithm in the centre and the estimates provided by the pKDTW-PWA algorithm on the right side.}
\label{fig:centroidsCBF}
\end{table*}

\subsection{KDTW-Centroid of a set of time series based on KDTW-PWA}

    \begin{algorithm}[H]
      \caption{pKDTW-PWA}\label{alg:pKDTW-PWA}
      \begin{algorithmic}[1]
        \Procedure{pKDTW-PWA}{$S$, $\nu$}
        \State //S: a set of time series of $D$ dimensional samples
        \State //$\nu$: the stiffness parameter of KDTW kernel 
        \State Ts $A$; //a D dimensional time series 
        \State SetOfTimeSeries $S_0$; 
        \While{$|S|>1$}
          \State $S_0= \emptyset$
          \While{$|S|>1$}
        		\State Let $ts_1$, $ts_2$ the first two time series in $S$;
        		\State Evaluate the AMA matrix for $ts_1$ and $ts_2$ 
        		\State \hspace{15mm}with $\nu$ as the stiffness parameter
        		\State $A$ = KDTW-PWA($ts_1$, $ts_2$, $AMA$);
        		\State $S_0=S_0 \cup \{A\}$;
        		\State $S=S\setminus\{ts_1, ts_2\}$;
        	  \EndWhile	
        	  \State $S=S_0 \cup S$;
        	\EndWhile
        	\State Let $A$ be the single element of $S$;
        \State \Return $A$
        \EndProcedure
      \end{algorithmic}
    \end{algorithm}

To average a larger set of time series using the pairwise average KDTW-PWA, we simply adopt the progressive agglomerative approach presented in Fig.\ref{fig:hac-iter-a}. This heuristic approach, detailed in Algorithm \ref{alg:pKDTW-PWA} has $O(n)$ complexity, $n$ being the size of the considered set of time series. 

The figures presented in Table \ref{fig:centroidsCBF} compare the centroid estimates provided by the iterated DBA, iKDBA and pKDTW-PWA algorithms. For the experiment, The DBA and iKDBA were iterated at most 20 times. Although the DBA and iKDBA estimates appear quite similar, the centroid estimates provided by the pKDTW-PWA algorithm is much smoother. This is a general property of the latter algorithm, which implements a time averaging principle based on the time expectation of sample occurrences, thus somehow allowing it to filter \textit{noisy} data.
Note also that the DBA and iKDBA estimates for the CBF data set are close to the results provided by the preimage approach (Fig.\ref{fig:preimageCBF}).

\section{Experimentation}

The purpose of this experiment is to evaluate the effectiveness of the proposed time elastic averaging methods against a double baseline, namely k-medoid-based approaches and the DBA algorithm. The first baseline allow us to compare centroid-based with medoid-based approaches. The second baseline highlights the advantages we can expect from using p.d elastic kernels instead of indefinite kernels such as DTW in the context of time series averaging. DBA is also currently considered as a state of the art method to average a set of sequences consistently with DTW.

For this purpose, we empirically evaluate the effectiveness of the methods using a first nearest centroid/medoid (1-NC) classification task on a set of time series derived from widely diverse fields of application. The task consists of representing each category contained in a training data set by estimating its medoid or centroid and then evaluating the error rate of a 1-NC classifier on an independent testing data set. Hence, the classification rule consists of assigning to the tested time series the category which corresponds to the closest (or most similar) medoid or centroid according to DTW or KDTW measures. \\

In \cite{Petitjean2014} a nice generalized k-NC task is described. The authors demonstrate that by selecting the appropriate number $k$ of centroids (using DBA and k-means), they achieve, without loss, a 70\% speed-up in average, compared to the original k-Near Neighbor task. Although, in general, the classification accuracies is improved when several centroids are used to represent the TRAIN datasets,  our main purpose is to highlight and amplify the  discrimination between time series averaging methods: this is why stick here with the 1-NC task. 

DBA and iKDBA iterative centroid methods are iterated at most 20 times and yield local estimates of the centroid. The pKDTW-PWA progressive agglomerative centroid method is only processed once, and hence is roughly 20 times faster than iKDBA and about 10 times faster than DBA.\\

A collection of 45 data sets is used to assess the proposed algorithms. The collection includes synthetic and real data sets, as well as univariate and multivariate time series. These data sets are distributed as follows: \\
\begin{itemize}
\item 42 of these data sets are available at the UCR repository \cite{KeoghUCRdataset}. Basically, we used all the data sets except for \textit{StarLightCurves}, \textit{Non-Invasive Fetal ECG Thorax1} and  \textit{Non-Invasive Fetal ECG Thorax2}. Although these last three data sets are still tractable, their computational cost is high because of their size and the length of the time series they contain. All the data sets are composed of scalar time series.
\item One data set, uWaveGestureLibrary\_3D was constructed from the uWaveGestureLibrary\_{X|Y|Z} scalar data sets to compose a new set of multivariate (3D) time series.
\item One data set, CharTrajTT, is available at the UCI Repository \cite{Lichman:2013} under the name \textit{Character Trajectories Data Set}. This data set contains multivariate (3D) time series and is divided into two equal sized data sets (TRAIN and TEST) for the experiment.  
\item The last data set, \textit{PWM2}, which stands for Pulse Width Modulation \cite{PWM}, was specifically defined to demonstrate a weakness in dynamic time warping (DTW) pseudo distance. This data set is composed of artificial scalar time series.\\
\end{itemize}
  
For each dataset, a training subset (TRAIN) is defined as well as an independent testing subset (TEST). We use the training sets to extract single medoids or centroid estimates for each of the categories defined in the data sets. 

Furthermore, for KDTW$_{Medoid}$, iKDBA and pKDTW-PWA, the $\nu$ parameter is optimized using a \textit{leave-one-out} (LOO) procedure carried out on the TRAIN data sets. The $\nu$ value is selected within the discrete set $\{.05, .1, .25, .5, 1, 2, 5, 10, 25, 50, 100\}$. The value that minimizes the LOO classification error rate on the TRAIN data is then used to provide the error rates that are estimated on the TEST data.\\

The classification results are given in Table \ref{tab:classResults}. It can be seen from this experiment, that 
\begin{enumerate}[i)]
\item Centroid-based methods outperform medoid-based methods: DBA yields lower error rates compared to DTW$_{Medoid}$, as do iKDBA and pKDTW-PWA compared to KDTW$_{Medoid}$.
\item iKDBA outperforms DBA: under the same experimental conditions (maximum of 20 iterations), the kernalized version of the DTW measure leads to better classification accuracy. To some extent, this confirms previous results obtained for SVM classification \cite{MarteauGibet2014} on such kinds of datasets.
\item pKDTW-PWA outperforms iKDBA: this results seems to show that joint averaging in the sample space and along the time axis improves the classification accuracy.  As pKDTW-PWA provides a centroid estimation in a single agglomerative step, we can conjecture that this method converges faster toward a satisfactory centroid candidate.\\ 
\end{enumerate}

The average ranking for all five tested methods, which supports our preliminary conclusion, is given at the bottom of Table \ref{tab:classResults}.\\

Following the study of \cite{Demsar:2006} on statistical tests available to evaluate the significance of differences in error rate between classifiers over multiple data sets, we conducted a Friedman's significance test, a sort of non-parametric counterpart of the well-known ANOVA. This test ranks the algorithms for each data set separately, the best performing algorithm being given a rank of 1, the second best rank 2, etc. 

\begin{figure*}[ht!]
\centering
\begin{tabular}{ccc}
	\includegraphics[scale=.35, angle=0]{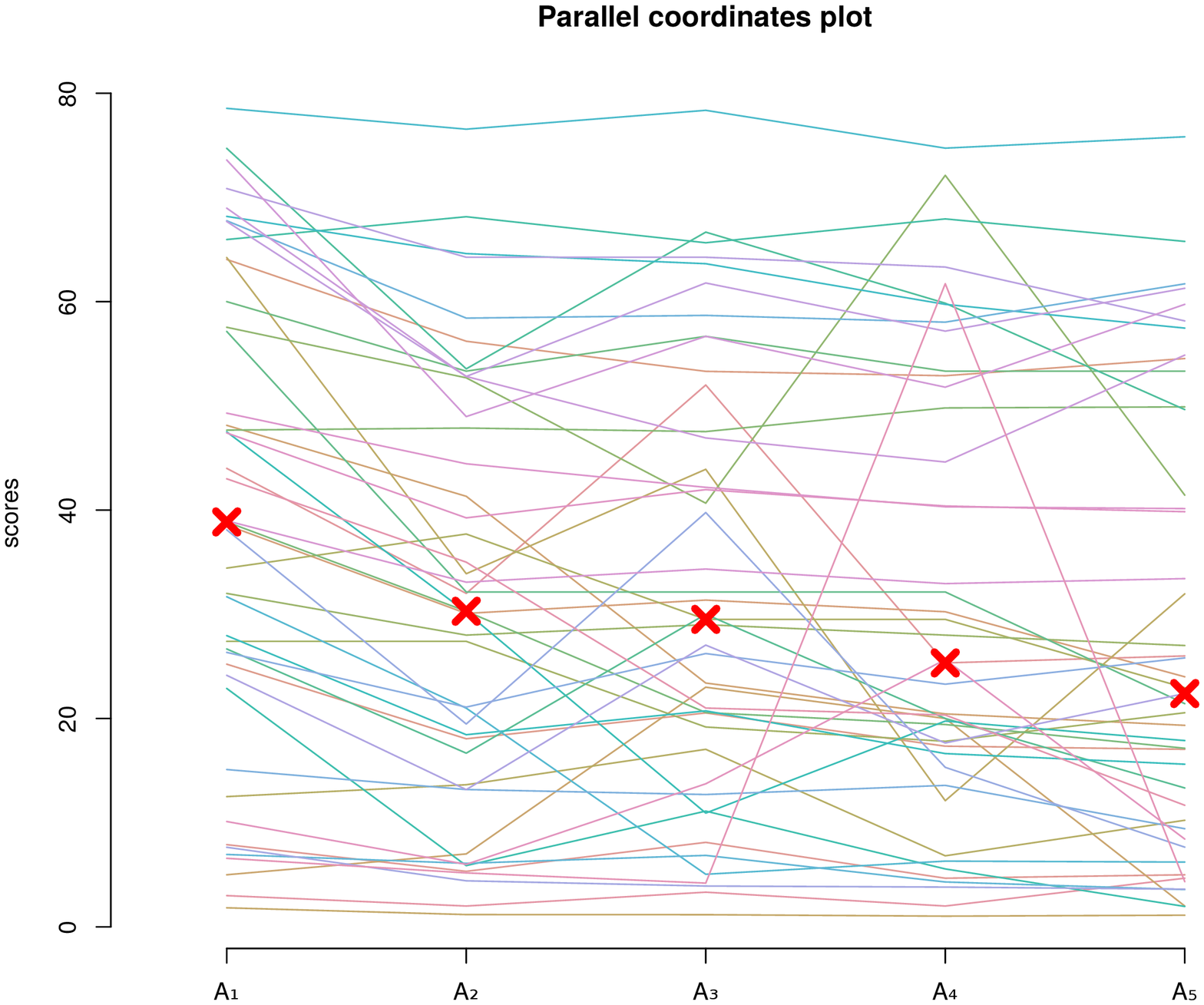} &
    	\includegraphics[scale=.35, angle=0]{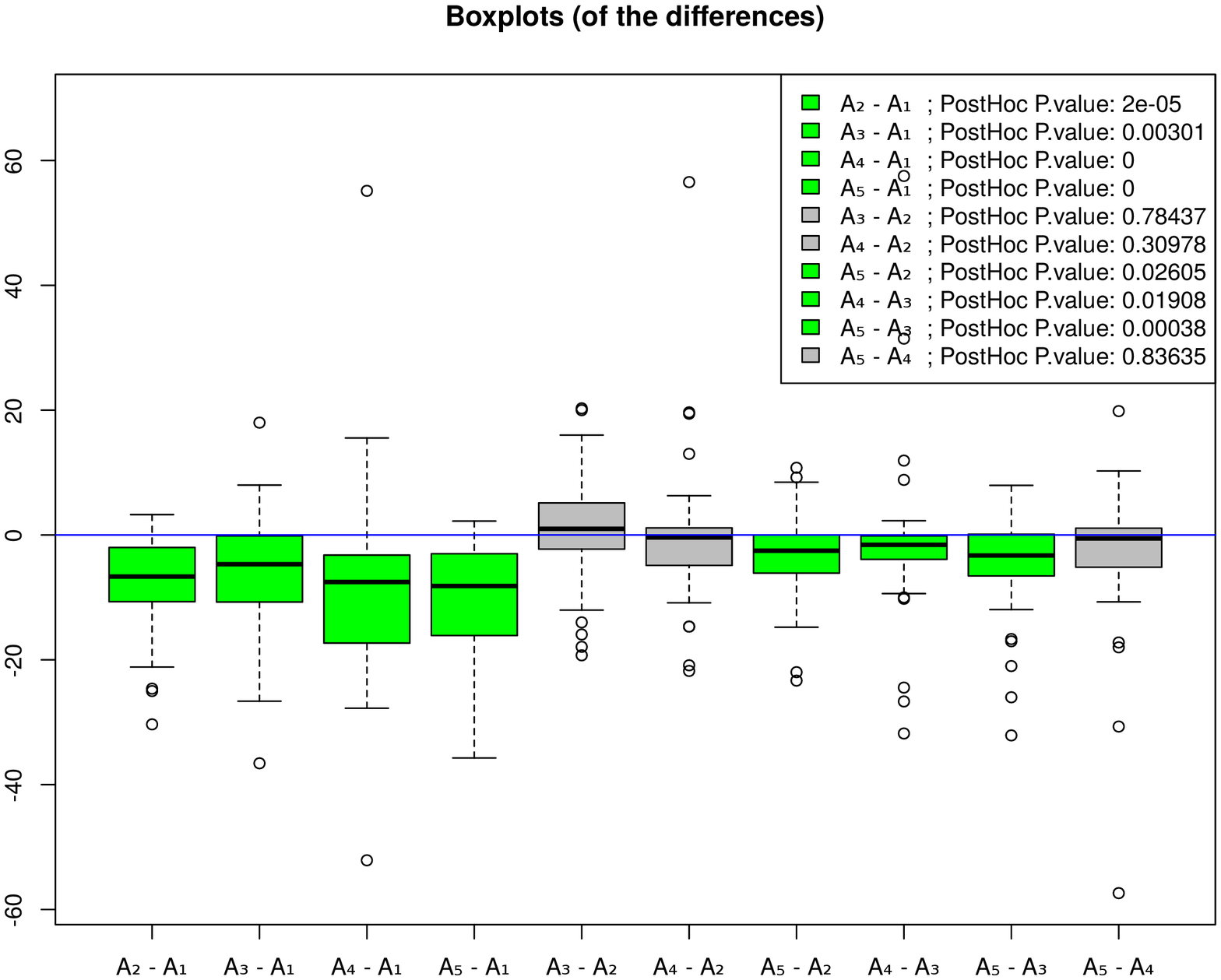} 
\end{tabular}
\caption{\textit{Post hoc} analysis of the Friedman's test: ($A_1$) DTW$_{Medoid}$, ($A_2$) DBA, ($A_3$) KDTW$_{Medoid}$, ($A_4$) iKDBA and ($A_5$) pKDTW-PWA.}
\label{fig:FriedmanTest}
\end{figure*}

According to this test, the null hypothesis is rejected (with a $P-value < 2.2e-16$). Post-hoc tests can then be carried out to compare pairwise algorithms using the Wilcoxon-Nemenyi-McDonald-Thompson test \cite{Hollander1999}. For this purpose, we use the R code provided by \cite{Galili:2010} to generate the parallel coordinate plots and boxplots presented in Fig.\ref{fig:FriedmanTest} as well as the results reported in Table \ref{tab:post-hoc}. \\

\begin{table*}[ht!]
\caption{Comparative study using the UCR and UCI data sets: classification error rates evaluated on the TEST data set (in \%) obtained using the first nearest neighbour classification rule for DTW$_{Medoid}$, DBA (centroid), KDTW$_{Medoid}$, $iKDBA$ (centroid) and $pKDTW-PWA$ (centroid). A single medoid/centroid extracted from the TRAIN data set represents each category.}
\label{tab:classResults}
\centering
\resizebox{\textwidth}{!}{\begin{tabular}{|l|c|c|c|c|c|c|}
\hline
\textbf{DATASET} & \# Cat $|$ L &  \textbf{DTW$_{Medoid}$} & \textbf{DBA} & \textbf{ KDTW$_{Medoid}$} & \textbf{iKDBA} & \textbf{pKDTW-PWA} \\
\hline\hline
Synthetic\_Control	& 6$|$60	& 3.00	&	\textbf{2.00}	&	3.33	&	\textbf{2.00}	&	4.67	\\
Gun\_Point & 2$|$150	&	44.00	&	32.00	&	52.00	&	\textbf{25.33}	&	\textbf{25.33}	\\
CBF & 3$|$128	&	7.89	&	5.33	&	8.11	&	\textbf{4.67}	&	5	\\
Face\_(all)	& 14$|$131 &	25.21	&	18.05	&	20.53	&	17.34	&	\textbf{17.04}	\\
OSU\_Leaf	& 6$|$427 &	64.05	&	56.20	&	53.31	&	\textbf{52.89}	&	54.54	\\
Swedish\_Leaf & 15$|$128	&	38.56	&	30.08	&	31.36	&	30.24	&	\textbf{24.00}	\\
50Words	&50$|$270 &	48.13	&	41.32	&	23.40	&	20.44	&	\textbf{19.34}	\\
Trace	&4$|$275&	5.00	&	7.00	&	23.00	&	20.00	&	\textbf{2.00}	\\
Two\_Patterns	&4$|$ 128  &	1.83	&	1.18	&	1.17	&	\textbf{1.03}	&	1.12	\\
Wafer	&2$|$152 &	64.23	&	33.89	&	43.92	&	\textbf{12.11}	&	31.96	\\
Face\_(four)	 &4$|$350 &	12.50	&	13.64	&	17.05	&	\textbf{6.82}	&	10.23	\\
Lightning-2	&2$|$637 &	34.43	&	37.70	&	29.51	&	29.51	&	\textbf{22.95}	\\
Lightning-7	&7$|$319 &	27.40	&	27.40	&	19.18	&	\textbf{17.81}	&	20.55	\\
ECG200	&2$|$96 &	32.00	&	28.00	&	29.00	&	28.00	&	\textbf{27.00}	\\
Adiac	&37$|$176 &	57.54	&	52.69	&	\textbf{40.67}	&	72.12	&	41.43	\\
Yoga		& 2$|$426&	47.67	&	47.87	&	\textbf{47.53}	&	49.80	&	49.90	\\
Fish		&7$|$463 &	38.86	&	30.29	&	20.57	&	19.42	&	\textbf{17.14}	\\
Beef		&5$|$470 &	60.00	&	\textbf{53.33}	&	56.67	&	\textbf{53.33}	&	\textbf{53.33}	\\
Coffee	&2$|$286 &	57.14	&	32.14	&	32.14	&	32.14	&	\textbf{21.43}	\\
OliveOil	 &4$|$570 &	26.67	&	16.67	&	30	&	20.00	&	\textbf{13.33}	\\
CinC\_ECG\_torso & 4$|$1639&	74.71	&	53.55	&	66.67	&	59.85	&	\textbf{49.64}	\\
ChlorineConcentration	&3$|$166 &	65.96	&	68.15	&	\textbf{65.65}	&	67.94	&	65.78	\\
DiatomSizeReduction	&4$|$345 &	22.88	&	5.88	&	11.11	&	5.56	&	\textbf{1.96}	\\
ECGFiveDays	&2$|$136 &	47.50	&	30.20	&	\textbf{10.92}	&	19.75	&	17.88	\\
FacesUCR	 &14$|$131 &	27.95	&	18.44	&	20.73	&	16.63	&	\textbf{15.61}	\\
Haptics	&5$|$1092 &	68.18	&	64.61	&	63.64	&	59.74	&	\textbf{57.47}	\\
InlineSkate	&7$|$1882 &	78.55	&	76.55	&	78.36	&	\textbf{74.73}	&	75.82	\\
ItalyPowerDemand &2$|$24	&	31.68	&	20.99	&	\textbf{5.05} 	&	6.31 	&	6.22	\\
MALLAT	&8$|$1024 &	6.95	&	6.10	&	6.87	&	4.22	&	\textbf{3.58}	\\
MedicalImages	&10$|$99 &	67.76	&	58.42	&	58.68	&	\textbf{58.03}	&	61.71	\\
MoteStrain	&2$|$84 &	15.10	&	13.18	&	12.70	&	13.58	&	\textbf{9.42}	\\
SonyAIBORobot\_SurfaceII	 &2$|$65 &	26.34	&	\textbf{21.09}	&	26.230	&	23.29	&	25.81	\\
SonyAIBORobot\_Surface	&2$|$70 &	38.10	&	19.47	&	39.77	&	15.31	&	\textbf{7.65}	\\
Symbols	&6$|$398 &	7.64	&	4.42	&	3.92	&	3.82	&	\textbf{3.62	}\\
TwoLeadECG	&2$|$82 &	24.14	&	\textbf{13.17}	&	27.04	&	17.65	&	22.39	\\
WordsSynonyms	&25$|$270 &	70.85	&	64.26	&	64.26	&	63.32	&	\textbf{58.15}	\\
Cricket\_X	&12$|$300 &	67.69	&	\textbf{52.82}	&	61.79	&	57.17	&	61.28	\\
Cricket\_Y	&12$|$300 &	68.97	&	52.82	&	46.92	&	\textbf{44.61}	&	54.87	\\
Cricket\_Z	&12$|$300 &	73.59	&	\textbf{48.97}	&	56.67	&	51.79	&	59.74	\\
uWaveGestureLibrary\_X	&8$|$315 &	38.97	&	33.08	&	34.34	&	\textbf{32.94}	&	33.42	\\
uWaveGestureLibrary\_Y	&8$|$315 &	49.30	&	44.44	&	42.18	&	40.31	&	\textbf{40.14}	\\
uWaveGestureLibrary\_Z	&8$|$315 &	47.40	&	\textbf{39.25}	&	41.96	&	40.39	&	39.84	\\
uWaveGestureLibrary\_3D	&8$|$315 &	10.11	&	\textbf{6.00}	&	13.74	&	25.65	&	8.43	\\
CharTrajTT\_3D	&20$|$178 &	6.58	&	5.18	&	\textbf{4.20} 	&	11.83	&	4.34	\\
PWM2	 & 3$|$128 &	43.00	&	35.00	&	21.00	&	20.33	&	\textbf{11.67}	\\
\hline
\hline
\textbf{\# Best Scores} & - & 0 & 8 & 6 & 13 & \textbf{22}   \\
\hline
\textbf{\# Uniquely Best Scores} & - & 0 & 6 & 3 & 10 & \textbf{20}   \\
\hline
\textbf{Average rank} & - & 4,29 & 2,8 & 3,16  & 2.22 & \textbf{1,89} \\
\hline
\end{tabular}}
\end{table*} 


\begin{table}[ht!]
 \caption{Significance test: $Algorithm_1$ is considered to be significantly better than $Algorithm_2$ according to the Friedman's test if the P-value (in bold characters) associated with the pairwise test is less than 0.05.}
\label{tab:post-hoc}
\centering
\begin{tabular}{|l|l|c|c|c|c|}
\hline
\textbf{$Algorithm_1$} & \textbf{$Algorithm_2$} & \textbf{P-value}\\
\hline\hline
DBA & DTW$_{Medoid}$ & \textbf{1.98e-05}\\
KDTW$_{Medoid}$ & DTW$_{Medoid}$ & \textbf{2.99e-03}\\
iKDBA & DTW$_{Medoid}$ & \textbf{1.38e-10}\\
pKDTW-PWA &  DTW$_{Medoid}$ & \textbf{1.09e-12}\\
KDTW$_{Medoid}$ & DBA & 7.84e-01\\
iKDBA & DBA & 3.10e-01\\
pKDTW-PWA & DBA & \textbf{2.60e-02}\\
iKDBA &  KDTW$_{Medoid}$ & \textbf{1.90e-02}\\
pKDTW-PWA & KDTW$_{Medoid}$ & \textbf{4.07e-04}\\
pKDTW-PWA & iKDBA & 8.36e-01\\
\hline
\end{tabular}
\end{table}

Table \ref{tab:post-hoc} reports the P-values for each pair of tested algorithms. This post-hoc analysis partially confirms our previous analysis of the classification results. If we consider that the null hypothesis is rejected when the P-value is less than $0.05$, the post-hoc analysis shows that centroid-based approaches perform significantly better than medoid-based approaches.  Furthermore, KDTW$_{Medoid}$ appears to be significantly better than DTW$_{Medoid}$. 

Furthermore, pKDTW-PWA is evaluated as significantly better than DBA but not significantly better than iKDBA in this experiment. Note also that DBA is not shown to perform significantly better than KDTW$_{Medoid}$.

\begin{figure}[ht!]
\centering
\begin{tabular}{ccc}
	\includegraphics[scale=.4, angle=0]{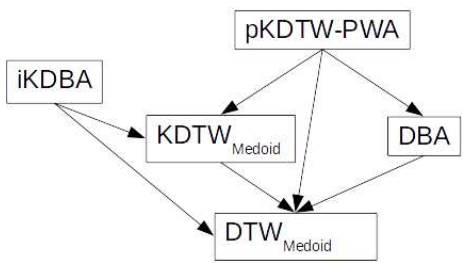}\\
\end{tabular}
\caption{Dominance graph for the five tested algorithms, according to the significance relation corresponding to Table \ref{tab:post-hoc} with a P-value threshold set at .05.}
\label{fig:SignificanceGraph}
\end{figure}

This post-hoc analysis is summarized in Fig.\ref{fig:SignificanceGraph} which shows the ranking graph for the five algorithms tested in our experiments.

\section{Conclusion}

In this paper, we address the reputedly difficult problem of averaging a set of time series in the context of a time elastic distance measure such as Dynamic Time Warping. The new perspective provided by the kernelization of the elastic distance firstly allows us to consider the averaging of time series as a preimage problem. This latter is unfortunately an ill-posed non-convex problem that could suffer from combinatorial number of local \textit{optima} when dealing with long multidimensional time series. Furthermore, this kind of preimage problem can only be resolved using gradient-free optimization procedures that are computationally very costly (since extensive functional evaluation is required).

However, this new kernelization approach allows a re-interpretation of pairwise kernel alignment matrices as distributions of probability over alignment paths. Based on this re-interpretation, we propose two distinct algorithms, iKDBA and pKDTW-PWA, based on iterative and progressive agglomerative heuristic methods, respectively, that are developed to compute approximate solutions to the multi-alignment of time series. 

We present an extensive experiment carried out on synthetic and real data sets, mostly containing univariate but also some multivariate time series. Our results show that centroid-based methods significantly outperform medoid-based methods in the context of a first nearest neighbour classification task. Most strikingly, the pKDTW-PWA algorithm, which integrates joint averaging in the sample space and along the time axis, is significantly better than the state-of-the art DBA algorithm, with a potentially lower computational cost. Indeed, the simple one-pass progressive agglomerative heuristic procedure is used in the pKDTW-PWA algorithm can be further optimized.



 \section*{Acknowledgments}
The authors thank the French Ministry of Research, the Brittany Region, the General Council of Morbihan and the European Regional Development Fund that partially funded this research. The authors also thank the promoters of the UCR and UCI data repositories for providing the time series data sets used in this study.

\bibliographystyle{IEEEtran}
\bibliography{biblio}   


\end{document}